# Operational Neural Networks

Serkan Kiranyaz[1], Turker Ince[2], Alexandros Iosifidis[3] and Moncef Gabbouj[4]

[1] Electrical Engineering, College of Engineering, Qatar University, Qatar; e-mail: mkiranyaz@qu.edu.qa
[2] Electrical & Electronics Engineering Department, Izmir University of Economics, Turkey; e-mail: turker.ince@izmirekonomi.edu.tr
[3] Department of Engineering, Aarhus University, Denmark; e-mail: alexandros.iosifidis@eng.au.dk
4 Department of Signal Processing, Tampere University of Technology, Finland; e-mail: Moncef.gabbouj@tut.fi

*Abstract—* Feed-forward, fully-connected Artificial Neural Networks (ANNs) or the so-called Multi-Layer Perceptrons (MLPs) are well-known universal approximators. However, their learning performance varies significantly depending on the function or the solution space that they attempt to approximate. This is mainly because of their homogenous configuration based solely on the linear neuron model. Therefore, while they learn very well those problems with a monotonous, relatively simple and linearly separable solution space, they may entirely fail to do so when the solution space is highly nonlinear and complex. Sharing the same linear neuron model with two additional constraints (local connections and weight sharing), this is also true for the conventional Convolutional Neural Networks (CNNs) and, it is, therefore, not surprising that in many challenging problems only the deep CNNs with a massive complexity and depth can achieve the required diversity and the learning performance. In order to address this drawback and also to accomplish a more generalized model over the convolutional neurons, this study proposes a novel network model, called Operational Neural Networks (ONNs), which can be heterogeneous and encapsulate neurons with any set of operators to boost diversity and to learn highly complex and multi-modal functions or spaces with minimal network complexity and training data. Finally, a novel training method is formulated to back-propagate the error through the operational layers of ONNs. Experimental results over highly challenging problems demonstrate the superior learning capabilities of ONNs even with few neurons and hidden layers.

## I. INTRODUCTION

The conventional fully-connected and feed-forward neural networks such as Multi-Layer Perceptrons (MLPs) and Radial Basis Functions (RBFs), are universal approximators. Such networks optimized by iterative processes [1][2], or even formed by random architectures and solving a closed-form optimization problem for the output weights [3], can approximate any continuous function providing that the employed neural units (i.e., the neurons) are capable of performing nonlinear piecewise continuous mappings of the receiving signals and that the capacity of the network (i.e. the number of layers' neurons) is sufficiently high. The standard approach in using such traditional neural networks is to manually define the network's architecture (i.e. the number of neural layers, the size of each layer) and use the same activation function for all neurons of the network.

While there is recently a lot of activity in searching for good network architectures based on the data at hand, either progressively [4], [5] or by following extremely laborious search strategies [6]-[10], the resulting network architectures may still exhibit a varying or entirely unsatisfactory performance levels especially when facing with highly complex and nonlinear problems. This is mainly due to the fact that all such traditional neural networks employ a homogenous network structure consisting of only a crude model of the biological neurons. This neuron model is capable of performing only the linear transformation (i.e., linear weighted sum) [12] while the biological neurons or neural systems in general are built from a large diversity of neuron types with heterogeneous, varying structural, biochemical and electrophysiological properties **[13]**-**[18]**. For instance, in mammalian retina there are roughly 55 different types of neurons to perform the low-level visual sensing **[16]**. Therefore, while these homogenous neural networks are able to approximate the responses of the training samples, they may not learn the actual underlying functional form of the mapping between the inputs and the outputs of the problem. There have been some attempts in the literature to modify MLPs by changing the neuron model and/or conventional BP algorithm [19]-[21], or the parameter updates [22], [23]; however, their performance improvements were not significant in general, since such approaches still inherit the main drawback of MLPs, i.e., homogenous network configuration with the same (linear) neuron model. Extensions of the MLP networks particularly for end-to-end learning of 2D (visual) signals, i.e. Convolutional Neural Networks (CNNs), and time-series data, i.e. Recurrent Neural Networks (RNNs) and Long Short-Term Memories (LSTMs), naturally inherit the same limitations originating from the traditional neuron model.

In biological learning systems, the limitations mentioned above are addressed at the neuron cell level [24]. In the mammalian brain and nervous system, each neuron (Figure 1) conducts the electrical signal over three distinct operations: 1) synaptic connections in *Dendrites*: an individual operation over each input signal from the synapse connection of the input neuron's axon terminals, 2) a pooling operation of the *operated* input signals via spatial and temporal signal integrator in the *Soma*, and finally, 3) an activation in the initial section of the *Axon* or the so-called *Axon hillock*: if the pooled potentials exceeds a certain limit, it "activates" a series of pulses (called action potentials). As shown in the right side of Figure 1 each terminal button is connected to other neurons across a small gap called *synapse*. The physical and neurochemical characteristics of each synapse determine the signal operation which is nonlinear in general [25], [26] along with the signal strength and polarity of the new input signal. Information storage or processing is concentrated in the cells' synaptic connections, or more precisely through certain operations of these connections together with the connection strengths (weights) [25]. Accordingly, in neurological systems, several distinct operations with proper weights (parameters) are created to accomplish such diversity and trained in time to perform or "to learn" many neural functions. Biological neural networks with higher diversity of computational operators have more computational power [28], and it is a fact that adding more neural diversity allows the network size and total connections to be reduced [24].

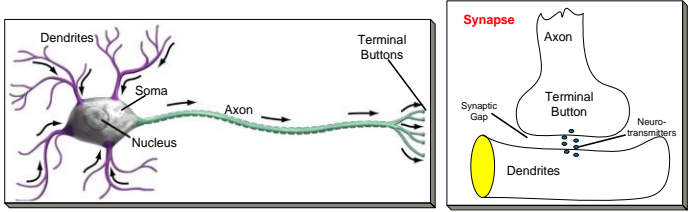

**Figure 1: A biological neuron (left) with the direction of the signal flow and a synapse (right).**

Motivated by these biological foundations, a novel feed-forward and fully-connected neural network model, called Generalized Operational Perceptrons (GOPs) [32], [33], has recently been proposed to accurately model the actual biological neuron with *varying* synaptic connections. In this heterogeneous configuration a superior diversity appearing in biological neurons and neural networks has been accomplished. More specifically, the diverse set of neurochemical operations in biological neurons (the non-linear synaptic connections plus the integration process occurring in the soma of a biological neuron model) have been modelled by the corresponding "Nodal" (synaptic connection) and "Pool" (integration in soma) operators whilst the "Activation" operator has directly been adopted. An illustrative comparison between the traditional Perceptron neuron in MLPs and the GOP neuron model is illustrated in Figure 2. Based on the fact that actual learning occurs in the synaptic connections with non-linear operators in general, those *all-time-fixed* linear model of MLPs can now be *generalized* by the GOP neurons that allow any (blend of) non-linear transformations to be used for defining the input signal transformations at the neuron level. Based on the fact that the GOP neuron naturally became a superset of linear Perceptrons (MLP neurons), GOPs provide an oportunity to better encode the input signal using linear and non-linear fusion schemes and, thus, lead to more compact neural network architectures achieving highly superior performance levels, e.g., the studies [32] and [33] have shown that GOPs can achieve elegant performance levels on many challenging problems where MLPs entirely fail to learn such as "Two-Spirals", "N-bit Parity" for N>10, "White Noise Regression", etc. As being the superset, a GOP network may fall back to a conventional MLP *only* when the learning process defining the neurons' operators indicates that the native MLP operators should be used for the learning problem in hand.

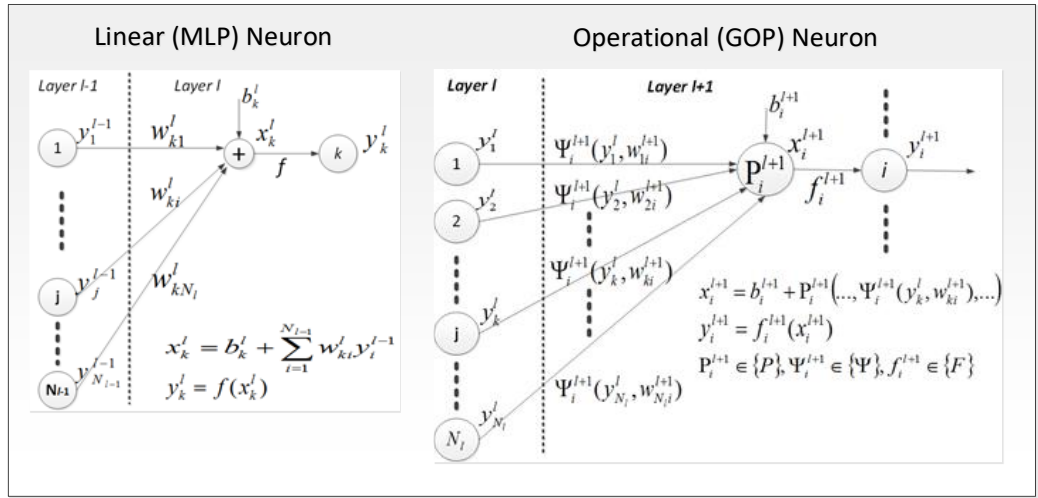

**Figure 2: Conventional MLP neuron (left) vs. GOP neuron with Nodal, $\Psi_i^{l+1}$,, Pool, $P_i^{l+1}$ , and Activation, $f_i^{l+1}$, operators.**

In this study, a novel neuron model is presented for the purpose of generalizing the linear neuron model of conventional CNNs with any non-linear operator. As an extension of the perceptrons, the neurons of a CNN perform the same linear transformation (i.e., the linear convolution, or equivalently, the linear weighted sum) as perceptrons do, and, it is, therefore, not suprising that in many challenging problems only the *deep* CNNs with a massive complexity and depth can achieve the required diversity and the learning performance. The main objective of this study is to propose a novel network model, the "Operational Neural Networks (ONNs)" based on this new neuron model. A novel training method is then formulated to back-propagate the error through the operational layers of ONNs. With the right operator set we shall show that ONNs even with a shallow and compact configuration and under severe restrictions (i.e., scarce and low-resolution train data, shallow training, limited operator library, etc.) can achieve an elegant learning performance over such challenging visual problems (e.g., image denoising, syntheses, transformation and segmentation) that can defy the conventional CNNs having the same or even higher network complexities. In order to perform an unbiased evaluation and direct comparison between the convolutional and operational neurons/layers, we shall avoid using the fully connected layers in both network types. This is a standard practice used by many state-of-the-art CNN topologies today.

The rest of the paper is organized as follows: In Section II, A brief review on GOPs is presented in order to highlight the motivation of a heterogeneous and nonlinear network, and to present how such a network can be trained by a modified BP. Based on the philosophy and foundations revealed in this section, Section III will then present the proposed ONNs and formulates the novel BP for training. Section IV presents a rich set of experiments to perform comparative evaluations between the learning performances of ONNs and CNNs over the four challenging problems. A detailed computational complexity analysis between the two network types will also be presented in this section. Finally, Section V concludes the paper and suggests topics for future research.

## II. PRELIMINARIES

### A. Generalized Operational Perceptrons (GOPs)

GOPs are the reference point for the proposed ONNs as they share the main philosophy of generalizing the conventional homogenous network only with the linear neuron model by a *heterogeneous* model with an "operational" neuron model which can encapsulate any set of (linear or non-linear) operators. As illustrated in Figure 2, the conventional feed-forward and fully-connected ANNs, or the so-called Multi-layer Perceptrons (MLPs), have the following linear model:

$$x_i^{l+1} = b_i^{l+1} + \sum_{k=1}^{N_l} y_k^l w_{ik}^{l+1}, \quad \forall i \in [1, N_{l+1}], \quad (1)$$

This means that the output of the previous layer neuron's output, $y_k^l$, contributes inputs of all neurons in the next layer, $l+1$. Then a nonlinear (or piece-wise linear) activation function is applied to all the neurons of layer $l+1$ in an element-wise manner. In a GOP neuron, this linear model has been replaced by an operator set of three operators: nodal operator, $\Psi_i^{l+1}$, pool operator, $P_i^{l+1}$ and finally the activation operator, $f_i^{l+1}$. The nodal operator models a synaptic connection with a certain neurochemical operation. The pool operator models the integration (or fusion) operation performed in Soma and finally, the activation operator encapsulates any activation function. Therefore, the output of the previous layer neuron, $y_k^l$, still contributes all the neurons' inputs in the next layer with the individual operator set of each neuron, i.e.,

$$x_i^{l+1} = b_i^{l+1} + P_i^{l+1}(\Psi_i^{l+1}(w_{1i}^{l+1}, y_1^l), \ldots, \Psi_i^{l+1}(w_{ki}^{l+1}, y_k^l), \ldots), \forall i \in [1, N_{l+1}] \quad (2)$$

Comparison of Eq. (2) with Eq. (1) reveals the fact that when $\Psi_i^{l+1}(w_{ki}^{l+1}, y_k^l) = w_{ki}^{l+1} \cdot y_k^l$ and $P_i^{l+1} = \sum(.)$ then the GOP neuron will be identical to a MLP neuron. However, in this relaxed model, now the neurons can get any proper nodal, pool and activation operator so as to maximize the learning capability. For instance, the nodal operator library, $\{\Psi\}$, can be composed of: *multiplication, exponential, harmonic (sinusoid), quadratic function, Gaussian, Derivative of Gaussian (DoG), Laplacian of Gaussian (LoG), Hermitian,* etc. Similarly, the pool operator library, $\{P\}$, can include: s*ummation, n-correlation, maximum, median*, etc. Typical activation functions that suit to classification problems can be combined within the activation operator library, $\{F\}$, composed of, e.g., *tanh, linear, lin-cut, binary,* etc. As in conventional MLP neuron, the $i$th GOP neuron at layer $l+1$ has the connection weights to each neuron in the previous layer, $l$; however, each weight is now the internal parameter of its nodal operator, $\Psi_i^{l+1}$, not necessarily the scalar weight of the output.

### B. Training with Back-Propagation

The conventional Back-Propagation (BP) training consists of one forward-propagation (FP) pass to compute the error at the output layer following with an error back-propagation pass starting from the output layer back to the 1st hidden layer, in order to calculate the individual weight and bias sensitivities in each neuron. The most common error metric is the Mean Square Error (MSE) in the output layer that can be expressed as follows:

$$E = E(y_1^L, \ldots, y_{N_L}^L) = \frac{1}{N_L} \sum_{i=1}^{N_L} (y_i^L - t_i)^2. \quad (3)$$

For an input vector $p$, and its corresponding output vector, $[y_1^L, \ldots, y_{N_L}^L]$, BP aims to compute the derivative of $E$ with respect to an individual weight, $w_{ik}^l$ (between the neuron $i$ and the output of the neuron $k$ in the previous layer, $l$-1), and bias, $b_i^l$, so that we can perform gradient descent method to minimize the error accordingly:

$$\frac{\partial E}{\partial w_{ik}^l} = \frac{\partial E}{\partial x_i^l} \frac{\partial x_i^l}{\partial w_{ik}^l} \quad and \quad \frac{\partial E}{\partial b_i^l} = \frac{\partial E}{\partial x_i^l} \frac{\partial x_i^l}{\partial b_i^l} = \frac{\partial E}{\partial x_i^l} \quad (4)$$

Both derivatives depend on the on the sensitivities of the error to the input, $x_i^l$. These sensitivities are usually called as *delta* errors. Let $\Delta_i^l = \partial E / \partial x_i^l$ be the delta error of the $i^{th}$ neuron at layer $l$. Now we can write the *delta* error by one step backward propagation from the output of that neuron, $y_i^l$ as follows:

$$\Delta_i^l = \frac{\partial E}{\partial x_i^l} = \frac{\partial E}{\partial y_i^l} \frac{\partial y_i^l}{\partial x_i^l} = \frac{\partial E}{\partial y_i^l} f'(x_i^l) \quad (5)$$

This means that the moment we found the derivative of the error to the output, $\partial E / \partial y_i^l$, we can then find the *delta* error. For the output layer, $l=L$, we know both terms:

$$\Delta_i^L = \frac{\partial E}{\partial x_i^L} = f'(x_i^L)(y_i^L - t_i) \quad (6)$$

Therefore, for both GOPs and MLPs, the delta error for each neuron at the output layer can be directly computed. One can also observe that Eqs. (4)-(6) are also common for both network types. However, the back-propagation of the delta error from the current layer (say $l+1$) to the previous layer, $l$, will be quite different. First consider for MLPs that Eq. (2) exhibits the contribution of the output of the $k^{th}$ neuron in the previous layer, $y_k^l$, to the input of each neurons of the current layer with individual weights, $w_{ik}^{l+1}$. With this in mind, one can express the derivative of the error to the output of the previous layer neuron, $\frac{\partial E}{\partial y_k^l}$, as follows:

$$\frac{\partial E}{\partial y_k^l} = \sum_{i=1}^{N_{l+1}} \frac{\partial E}{\partial x_i^{l+1}} \frac{\partial x_i^{l+1}}{\partial y_k^l} = \sum_{i=1}^{N_{l+1}} \Delta_i^{l+1} w_{ik}^{l+1} \quad (7)$$

Now one can use Eq. (5) to lead to a generic equation of the back-propagation of the delta errors, as follows:

$$\Delta_k^l = \frac{\partial E}{\partial x_k^l} = \frac{\partial E}{\partial y_k^l} f'(x_k^l) = f'(x_k^l) \sum_{i=1}^{N_{l+1}} \Delta_i^{l+1} w_{ik}^{l+1} \quad (8)$$

So, for MLPs another linear transformation with the same weights are used to back-propagate the delta errors of the current layer to compute the delta errors of the previous layer. For GOPs, this turns out to be a different scheme. From the Eq. (2) the same output derivative, $\partial E / \partial y_k^l$, can be expressed as follows:

$$\frac{\partial E}{\partial y_k^l} = \sum_{i=1}^{N_{l+1}} \frac{\partial E}{\partial x_i^{l+1}} \frac{\partial x_i^{l+1}}{\partial y_k^l} = \sum_{i=1}^{N_{l+1}} \Delta_i^{l+1} \frac{\partial x_i^{l+1}}{\partial P_i^{l+1}} \frac{\partial P_i^{l+1}}{\partial \Psi_i^{l+1}(w_{ki}^{l+1}, y_k^l)} \frac{\partial \Psi_i^{l+1}(w_{ki}^{l+1}, y_k^l)}{\partial y_k^l} \quad (9)$$

where $\partial x_i^{l+1}/\partial P_i^{l+1} = 1$. Let $\nabla_{\Psi_{ki}} P_i^{l+1} = \frac{\partial P_i^{l+1}}{\partial \Psi_i^{l+1}(w_{ki}^{l+1}, y_k^l)}$ and, $\nabla_y \Psi_{ki}^{l+1} = \frac{\partial \Psi_i^{l+1}(w_{ki}^{l+1}, y_k^l)}{\partial y_k^l}$. Then Eq. (9) becomes:

$$\frac{\partial E}{\partial y_k^l} = \sum_{i=1}^{N_{l+1}} \Delta_i^{l+1} \nabla_{\Psi_{ki}} P_i^{l+1} \nabla_y \Psi_{ki}^{l+1} \qquad (10)$$

Obviously both $\nabla_{\Psi_{ki}} P_i^{l+1}$ and $\nabla_y \Psi_{ki}^{l+1}$ will be different functions for different nodal and pool operators. From the output sensitivity, $\partial E/\partial y_k^l$, one can get the delta of that neuron, $\Delta_k^l$, which leads to the generic equation of the back-propagation of the delta errors for GOPs, as follows:

$$\Delta_k^l = \frac{\partial E}{\partial x_k^l} = \frac{\partial E}{\partial y_k^l} f'(x_k^l) = \\ f'(x_k^l) \sum_{i=1}^{N_{l+1}} \Delta_i^{l+1} \nabla_{\Psi_{ki}} P_i^{l+1} \nabla_y \Psi_{ki}^{l+1} \qquad (11)$$

Once all the delta errors in each layer are formed by back-propagation, then weights and bias of each neuron can be updated by the gradient descent method. Note that a bias sensitivity in GOPs is identical as MLPs,

$$\frac{\partial E}{\partial b_k^l} = \Delta_k^l \qquad (12)$$

For the weight sensitivity, one can express the chain rule of derivatives as,

$$\frac{\partial E}{\partial w_{ki}^{l+1}} = \frac{\partial E}{\partial x_i^{l+1}} \frac{\partial x_i^{l+1}}{\partial w_{ki}^{l+1}} = \\ \Delta_i^{l+1} \frac{\partial x_i^{l+1}}{\partial P_i^{l+1}} \frac{\partial P_i^{l+1}}{\partial \Psi_i^{l+1}(w_{ki}^{l+1}, y_k^l)} \frac{\partial \Psi_i^{l+1}(w_{ki}^{l+1}, y_k^l)}{\partial w_{ki}^{l+1}} \qquad (13)$$

where $\partial x_i^{l+1}/\partial P_i^{l+1} = 1$. Let $\nabla_w \Psi_{ki}^{l+1} = \frac{\partial \Psi_i^{l+1}(w_{ki}^{l+1}, y_k^l)}{\partial w_{ki}^{l+1}}$. Then Eq. (13) simplifies to,

$$\frac{\partial E}{\partial w_{ki}^{l+1}} = \Delta_i^{l+1} \nabla_{\Psi_{ki}} P_i^{l+1} \nabla_w \Psi_{ki}^{l+1} \qquad (14)$$

For different nodal operators along with their derivatives, $\nabla_w \Psi_{ki}^{l+1}$ and $\nabla_y \Psi_{ki}^{l+1}$ with respect to the weight, $w_{ki}^{l+1}$, and the output, $y_k^l$ of the previous layer neurons, Eqs. (12) and (14) will yield the weight and bias sensitivities. Therefore, the operator set of each neuron should be assigned before the application of BP training. However, this is a typical "Chicken and Egg" problem because finding out the right operators even for a single neuron eventually requires a trained network to evaluate the learning performance. Furthermore, the optimality of the operator set of that neuron obviously depends on the operators of the other neurons since variations in the latter can drastically change the optimality of the earlier operator choice for that neuron. The greedy iterative search (GIS) was proposed in [32] and [33]. The basic idea of GIS is to reduce the search space significantly so that the layer-wise pruned search finds near-optimal operator sets per layer (for all neurons in that layer).

## III. OPERATIONAL NEURAL NETWORKS

The convolutional layers of conventional 2D CNNs share the same neuron model as in MLPs with two additional restrictions: limited connections and weight sharing. Without these restrictions every pixel in a feature map in a layer would be connected to every pixel of a feature map at the previous layer and this would create an infeasibly large number of connections and weights that cannot be optimized efficiently. Instead, by these two constraints a pixel in the current layer will now be connected only to the corresponding neighboring pixels in the previous layer (limited connections) and the amount of connections can be determined by the size of the kernel (filter). Moreover, the connection weights of the kernel will be shared for each pixel-to-pixel connection (weight sharing). By these restrictions, the linear weighted sum as expressed in Eq. (1) for MLPs will turn into the convolution formula used in CNNs. This is also evident in the illustration in Figure 3 (left) where the three consecutive convolutional layers without the sub-sampling (pooling) layers are shown. So, the input map of the next layer neuron, $x_k^l$, will be obtained by cumulating the final output maps, $y_i^{l-1}$ of the previous layer neurons convolved with their individual kernels, $w_{ki}^l$, as follows:

$$x_k^l = b_k^l + \sum_{i=1}^{N_{l-1}} conv2D(w_{ki}^l, y_i^{l-1}, 'NoZeroPad') \\ \therefore x_k^l(m,n)\Big|_{(0,0)}^{(M-1,N-1)} = \\ \sum_{r=0}^{2} \sum_{t=0}^{2} \left(w_{ki}^l(r,t) y_i^{l-1}(m+r, n+t)\right) + \cdots \qquad (15)$$

ONNs share the essential idea of GOPs and extends the sole usage of linear convolutions in the convolutional layers of CNNs by the nodal and pool operators. In this way, the operational layers and neurons constitute the backbone of an ONN and other properties such as weight sharing and limited (kernel-wise) connectivity are common with a CNN. The three consecutive operational layers and the $k^{th}$ neuron of the sample ONN with 3x3 kernels and $M=N=22$ input map sizes in the previous layer is shown in Figure 3 (right). The input map of the $k^{th}$ neuron at the current layer, $x_k^l$, is obtained by *pooling* the final output maps, $y_i^{l-1}$ of the previous layer neurons *operated* with its corresponding kernels, $w_{ki}^l$, as follows:

$$x_k^l = b_k^l + \sum_{i=1}^{N_{l-1}} oper2D(w_{ki}^l, y_i^{l-1}, 'NoZeroPad') \\ x_k^l(m,n)\Big|_{(0,0)}^{(M-1,N-1)} = b_k^l + \\ \sum_{i=1}^{N_{l-1}} \left(P_k^l \begin{bmatrix} \Psi_{ki}^l\left(w_{ki}^l(0,0), y_i^{l-1}(m,n)\right), \dots, \\ \Psi_{ki}^l\left(w_{ki}^l(r,t), y_i^{l-1}(m+r, n+t), \dots\right), \dots \end{bmatrix}\right) \qquad (16)$$

A direct comparison between Eqs. (15) and (16) will reveal the fact that when the pool operator is "summation", $P_k^l = \Sigma$, and the nodal operator is "multiplication", $\Psi_{ki}^l(y_i^{l-1}(m,n), w_{ki}^l(r,t)) = w_{ki}^l(r,t) y_i^{l-1}(m,n)$, for all neurons of an ONN, then the resulting homogenous ONN will be identical to a CNN. Therefore, as the GOPs are the superset of MLPs, ONNs are a superset of CNNs.

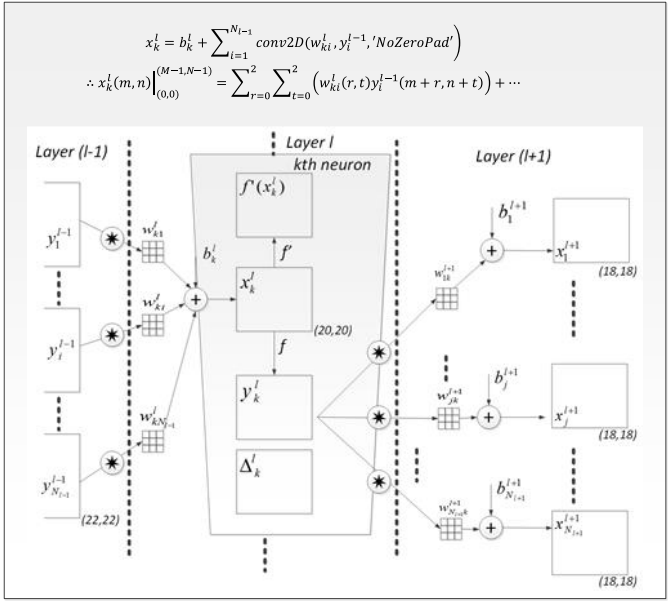 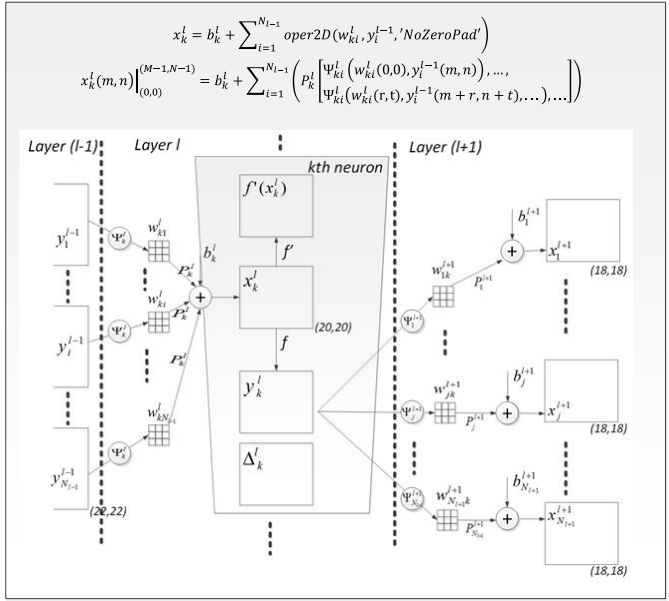

**Figure 3: Three consecutive convolutional (left) and operational (right) layers with the $k^{th}$ neuron of a CNN (left) and an ONN (right).**

### A. Training with Back-Propagation

The formulation of BP training consists of four distinct phases: 1) Computation of the delta error, $\Delta_1^L$, at the output layer, 2) Inter BP between two operational layers, 3) Intra BP in an operational neuron, and 4) Computation of the weight (operator kernel) and bias sensitivities in order to update them at each BP iteration. Phase-3 also takes care of up- or down-sampling (pooling) operations whenever they are applied in the neuron. In order to explain each phase in detail, we shall first formulate its counterpart for the convolutional layers (and neurons) of a conventional CNN and then we will present the corresponding formulation for the operational layers (and neurons) while highlighting the main differences. In this way, the BP analogy between MLPs and GOPs presented in Section II.B will be constructed this time between CNNs and ONNs. One can also witness how the BP formulation for GOPs will alter for ONNs due to the two aforementioned restrictions. For this section, we shall assume a particular application, e.g., object segmentation, to exemplify a learning objective for training.

#### 1) BP from the output ONN layer

This phase is common for ONNs and CNNs. As shown in Figure 4, the output layer has only one output neuron from which the initial delta error (the sensitivity of the input with respect to the object segmentation error) is computed. In the most basic terms the object segmentation error for an image $I$ in the dataset can be expressed as the Mean-Square-Error (MSE) between the object's segmentation mask (SM) and the real output, $y_1^L$.

$$E(I) = \sum_p \left(y_1^L(I_p) - SM(I_p)\right)^2 \quad (17)$$

where $I_p$ is the pixel $p$ of the image $I$. The delta sensitivity of the error can then be computed as:

$$\Delta_1^L = \frac{\partial E}{\partial x_1^L} = \frac{\partial E}{\partial y_1^L}\frac{\partial y_1^L}{\partial x_1^L} = (y_1^L(I) - SM(I))f'(x_1^L(I)) \quad (18)$$

Note that the delta error is proportional to the difference between the real output and the segmentation mask. For ONNs any differentiable error function can be used besides MSE.

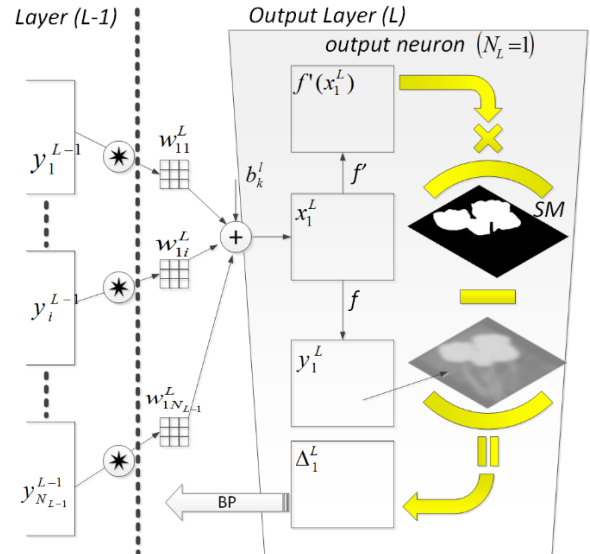

**Figure 4: Delta error computation at the output layer for segmentation.**

#### 2) Inter-BP among ONN layers: $\Delta y_k^l \overset{\Sigma}{\leftarrow} \Delta_i^{l+1}$

Once the delta error is computed in the output layer, it will be back-propagated to the hidden layers of the network. The basic rule of BP states the following: *if the output of the $k^{th}$ neuron at layer l, contributes a neuron i in the next level with a weight, $w_{ki}^l$, that next layer neuron's delta, $\Delta_i^{l+1}$, will contribute with the same* weight to form $\Delta_k^l$ of the neuron in the previous layer, $l$. This can be expressed as,

$$\frac{\partial E}{\partial y_k^l} = \Delta y_k^l \overset{BP}{\leftarrow} \sum \Delta_1^{l+1}, \forall i \in \{1, N_{l+1}\} \quad (19)$$

Specifically:

$$\frac{\partial E}{\partial y_k^l} = \Delta y_k^l = \sum_{i=1}^{N_{l+1}} \frac{\partial E}{\partial x_i^{l+1}} \frac{\partial x_i^{l+1}}{\partial y_k^l} = \sum_{i=1}^{N_{l+1}} \Delta_1^{l+1} \frac{\partial x_i^{l+1}}{\partial y_k^l} \quad (20)$$

where the delta error in the next (output) layer is already computed; however, the derivative, $\frac{\partial x_i^{l+1}}{\partial y_k^l}$, differs between a convolutional and operational layer. In the former, the input-output expression is,

$$x_i^{l+1} = \ldots + y_k^l * w_{ik}^{l+1} + \ldots \quad (21)$$

It is obviously hard to compute the derivative directly from the convolution. Instead we can focus on a single pixel's contribution of the output, $y_k^l(m,n)$, to the pixels of the $x_i^{l+1}(m,n)$ is shown. Assuming a 3x3 kernel, Eq. (22) presents the contribution of the $y_k^l(m,n)$ to the 9 neighbor pixels. This is illustrated on Figure 5 where the role of an output pixel, $y_k^l(m,n)$, over two pixels of the next layer's input neuron's pixels, $x_i^{l+1}(m-1,n-1)$ and $x_i^{l+1}(m+1,n+1)$. Considering the pixel as a MLP neuron that are connected to other MLP neurons in the next layer, according to the basic rule of BP one can express the delta of $y_k^l(m,n)$ as in Eq. (23).

$$\begin{aligned}
x_i^{l+1}(m-1,n-1) &= \ldots + y_k^l(m,n). w_{ik}^{l+1}(2,2) + \ldots \\
x_i^{l+1}(m-1,n) &= \ldots + y_k^l(m,n). w_{ik}^{l+1}(2,1) + \ldots \\
x_i^{l+1}(m,n) &= \ldots + y_k^l(m,n). w_{ik}^{l+1}(1,1) + \ldots \\
&\ldots \\
x_i^{l+1}(m+1,n+1) &= \ldots + y_k^l(m,n). w_{ik}^{l+1}(0,0) + \ldots \\
\therefore x_i^{l+1}(m+r,n+t)\Big|_{(1,1)}^{(M-1,N-1)} &= w_{ik}^{l+1}(1-r,1-t)\, y_k^l(m,n) + ..
\end{aligned} \quad (22)$$

$$\begin{aligned}
\frac{\partial E}{\partial y_k^l}(m,n)\Big|_{(0,0)}^{(M-1,N-1)} &= \Delta y_k^l(m,n) = \sum_{i=1}^{N_{l+1}} \left( \sum_{r=-1}^{1} \sum_{t=-1}^{1} \frac{\partial E}{\partial x_1^{l+1}(m+r,n+t)} \frac{\partial x_1^{l+1}(m+r,n+t)}{\partial y_k^l(m,n)} \right) \\
&= \sum_{i=1}^{N_{l+1}} \left( \sum_{r=-1}^{1} \sum_{t=-1}^{1} \Delta_1^{l+1}(m+r,n+t). w_{ik}^{l+1}(1-r,1-t) \right)
\end{aligned} \quad (23)$$

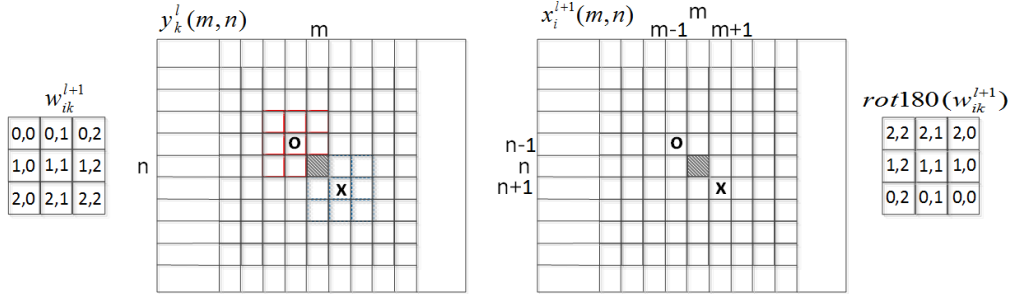

**Figure 5: A single pixel's contribution of the output, $y_k^l(m,n)$, to the two pixels of the $x_i^{l+1}$ using a 3x3 kernel.**

If we generalize Eq. (23) for all pixels of the $\Delta y_k^l$:

$$\Delta y_k^l = \sum_{i=1}^{N_{l+1}} conv2D(\Delta_i^{l+1}, rot180(w_{ik}^{l+1}), 'ZeroPad') \quad (24)$$

Interestingly, this expression also turns out to be a *full* convolution with zero padding by $(Kx-1, Ky-1)$ zeros to each boundary of the $\Delta_i^{l+1}$ in order to obtain equal dimensions (width and height) for $\Delta y_k^l$ with $y_k^l$.

In an operational layer, the input-output expression becomes,

$$x_i^{l+1} = \ldots + oper2D(y_k^l, w_{ik}^{l+1}) + \ldots \quad (25)$$

where $oper2D(y_k^l, w_{ik}^{l+1})$ represents a 2D operation with a particular pool and nodal operator. Once again, it is not feasible to compute the derivative directly from this 2D operation formula (pooling of the outputs of the nodal operators). Instead we can again focus on a single pixel's contribution of the output, $y_k^l(m,n)$, to the pixels of the $x_i^{l+1}(m,n)$. Assuming again a $Kx \times Ky = 3 \times 3$ kernel, Eq. (26) formulates the contribution of the $y_k^l(m,n)$ to the 9 neighbor pixels. This is illustrated on Figure 6 where the contribution of an output pixel, $y_k^l(m,n)$, over the two pixels of the next layer's input neuron's pixels, $x_i^{l+1}(m-1,n-1)$ and $x_i^{l+1}(m-2,n-2)$ is shown. Considering the pixel as a "GOP neuron" that is connected to other GOP neurons in the next layer, according to the basic rule of BP one can then formulate the delta of $y_k^l(m,n)$ as in Eq. (27). Note that this is slightly different than what we derived for the contribution of the $y_k^l(m,n)$ for convolutional layers, expressed in in Eq. (23) and illustrated in Figure 5. In that case the output pixel, $y_k^l(m,n)$, and input pixel, $x_i^{l+1}(m,n)$, were connected through the center of the kernel, i.e., $x_i^{l+1}(m,n) = \ldots + y_k^l(m,n). w_{ik}^{l+1}(1,1) + \ldots$ This connection has led to the rotation (by 180 degrees) operation for the BP of the delta error. In order to avoid this undesired rotation, the output pixel, $y_k^l(m,n)$, and input pixel, $x_i^{l+1}(m,n)$, are connected through the first (top-left) element of the kernel, i.e., $x_i^{l+1}(m,n) = \ldots + P_i^{l+1}[\Psi_i^{l+1}(y_k^l(m,n), w_{ik}^{l+1}(0,0)), \ldots, \Psi_i^{l+1}(y_k^l(m+r,n+t), w_{ik}^{l+1}(r,t),)\ldots)]$. This means that the contribution of the output pixel, $y_k^l(m,n)$, will now only be on the $x_i^{l+1}(m-r, n-t)$ as expressed in Eq. (26). The major difference over the delta-error BP expressed in Eqs. (23) and (24) is that the chain-rule of derivatives should now include the two operator functions, pool and nodal, both of which were fixed to summation and multiplication before. The delta error of the output pixel can, therefore, be expressed as in Eq. (27) in the generic form of pool, $P_i^{l+1}$, and nodal, $\Psi_i^{l+1}$, operator functions of each operational neuron $i \in [1, \ldots, N_{l+1}]$ in the next layer.

$$\begin{aligned}
x_i^{l+1}(m-1, n-1) &= \ldots + P_i^{l+1}[\Psi_i^{l+1}(y_k^l(m-1,n-1), w_{ik}^{l+1}(0,0)), \ldots, \Psi_i^{l+1}(y_k^l(m,n), w_{ik}^{l+1}(1,1))] + \ldots \\
x_i^{l+1}(m-1, n) &= \ldots + P_i^{l+1}[\Psi_i^{l+1}(y_k^l(m-1,n), w_{ik}^{l+1}(0,0)), \ldots, \Psi_i^{l+1}(y_k^l(m,n), w_{ik}^{l+1}(1,0)), \ldots] + \ldots \\
x_i^{l+1}(m, n) &= \ldots + P_i^{l+1}[\Psi_i^{l+1}(y_k^l(m,n), w_{ik}^{l+1}(0,0)), \ldots, \Psi_i^{l+1}(y_k^l(m+r, n+t), w_{ik}^{l+1}(r,t),)\ldots)] + \ldots \\
&\ldots \ldots \\
x_i^{l+1}(m+1, n+1) &= \ldots + P_i^{l+1}[\Psi_i^{l+1}(y_k^l(m+1, n+1), w_{ik}^{l+1}(0,0)), \ldots] + \ldots \\
\therefore x_i^{l+1}(m-r, n-t)\Big|_{(1,1)}^{(M-1,N-1)} &= b_i^{l+1} + \sum_{k=1}^{N_1} P_i^{l+1}\big[\ldots, \Psi_i^{l+1}(w_{ik}^{l+1}(r,t), y_k^l(m,n)), \ldots\big]
\end{aligned} \quad (26)$$

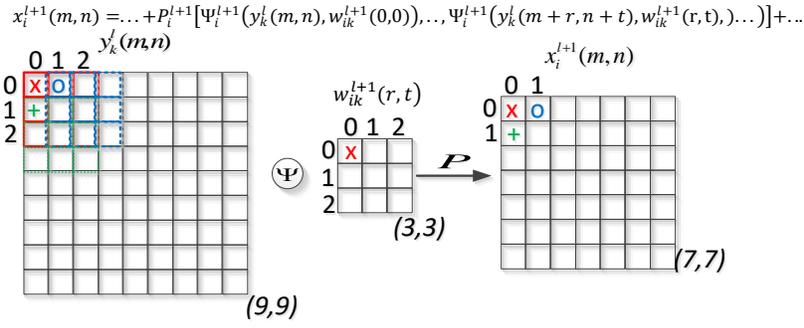

**Figure 6: Starting from (0,0), a single pixel's contribution of the output, $y_k^l(m,n)$, to the two pixels of the $x_i^{l+1}$ using a $Kx \times Ky = 3 \times 3$ kernel.**

$$\therefore \frac{\partial E}{\partial y_k^l}(m,n)\Big|_{(0,0)}^{(M-1,N-1)} = \Delta y_k^l(m,n) = \sum_{i=1}^{N_{l+1}} \left( \sum_{r=0}^{Kx-1} \sum_{t=0}^{Ky-1} \frac{\frac{\partial E}{\partial x_i^{l+1}(m-r,n-t)} \times \frac{\partial x_i^{l+1}(m-r,n-t)}{\partial P_i^{l+1}[..,\Psi_i^{l+1}(y_k^l(m,n), w_{ik}^{l+1}(r,t)),..]} \times}{\frac{\partial P_i^{l+1}[..,\Psi_i^{l+1}(y_k^l(m,n), w_{ik}^{l+1}(r,t)),..]}{\partial \Psi_i^{l+1}(y_k^l(m,n), w_{ik}^{l+1}(r,t))} \times \frac{\partial \Psi_i^{l+1}(y_k^l(m,n), w_{ik}^{l+1}(r,t))}{\partial y_k^l(m,n)}} \right) \quad (27)$$

In Eq. (27), note that the first term, $\frac{\partial x_i^{l+1}(m-r,n-t)}{\partial P_i^{l+1}[..,\Psi_i^{l+1}(y_k^l(m,n), w_{ik}^{l+1}(r,t)),..]} = 1$. Let

$\nabla_{\Psi_{ki}} P_i^{l+1}(m,n,r,t) = \frac{\partial P_i^{l+1}[..,\Psi_i^{l+1}(y_k^l(m,n), w_{ik}^{l+1}(r,t)),..]}{\partial \Psi_i^{l+1}(y_k^l(m,n), w_{ik}^{l+1}(r,t))}$ and

$\nabla_y \Psi_{ki}^{l+1}(m,n,r,t) = \frac{\partial \Psi_i^{l+1}(y_k^l(m,n), w_{ik}^{l+1}(r,t))}{\partial y_k^l(m,n)}$. First, it is obvious that both derivatives, $\nabla_{\Psi_{ki}} P_i^{l+1}$, and $\nabla_y \Psi_{ki}^{l+1}$, no longer require the rotation of the kernel, $w_{ik}^{l+1}$. The first derivative, $\nabla_{\Psi_{ki}} P_i^{l+1}$, depends on the role (contribution) of the particular nodal term, $\Psi_i^{l+1}(y_k^l(m,n), w_{ik}^{l+1}(r,t))$, within the pool function. The derivative, $\nabla_{\Psi_{ki}} P_i^{l+1}(m,n,r,t)$ is computed while computing the pixels $x_i^{l+1}(m-r, n-t)$ for $\forall r, t \in (Kx, Ky)$ that corresponds to the particular output value, $y_k^l(m,n)$, within each pool function. Recall that this is the contribution of the $y_k^l(m,n)$ alone for each input value at the next layer, $x_i^{l+1}(m-r, n-t)$ for $\forall r, t \in (Kx, Ky)$. When the pool operator is summation, $P_i^{l+1} = \Sigma$, then $\nabla_{\Psi_{ki}} P_i^{l+1} = 1$, which is constant for any nodal term. For any other alternative, the derivative $\nabla_{\Psi_{ki}} P_i^{l+1}(m,n,r,t)$ will be a function of four variables. The second derivative, $\nabla_y \Psi_{ki}^{l+1}$, is the derivative of the nodal operator with respect to the output. For instance, when the nodal operator is the common operator of the convolutional neuron, "multiplication", i.e., $\Psi_i^{l+1}(y_k^l(m,n), w_{ik}^{l+1}(r,t)) = y_k^l(m,n) \cdot w_{ik}^{l+1}(r,t)$, then this derivative is simply the weight kernel, $w_{ik}^{l+1}(r,t)$. This is the only case where this derivative will be independent from the output, $y_k^l(m,n)$. For any other alternative, the derivative $\nabla_y \Psi_{ki}^{l+1}(m,n,r,t)$ will also be a function of four variables. By using these four variable derivatives or equivalently two 4-D matrices, Eq. (27) can be simplified as Eq. (28). Note that $\Delta y_k^l$, $\nabla_{\Psi_{ki}} P_i^{l+1}$ and $\nabla_y \Psi_{ki}^{l+1}$ have the size, $M \times N$ while the next layer delta error, $\Delta_i^{l+1}$, has the size, $(M - K_x + 1) \times (N - K_y + 1)$, respectively. Therefore, to enable this variable 2D convolution in this equation, the delta error, $\Delta_i^{l+1}$, is padded zeros on all the four boundaries ($K_x - 1$ zeros on left and right, $K_y - 1$ zeros on the bottom and top).

$$\Delta y_k^l(m,n)\Big|_{(0,0)}^{(M-1,N-1)} = \sum_{i=1}^{N_{l+1}} \left( \sum_{r=0}^{K_x-1} \sum_{t=0}^{K_y-1} \Delta_i^{l+1}(m-r,n-t) \times \nabla_{\Psi_{ki}} P_i^{l+1}(m,n,r,t) \times \nabla_y \Psi_{ki}^{l+1}(m,n,r,t) \right)$$

$$\text{Let } \nabla_y P_i^{l+1}(m,n,r,t) = \nabla_{\Psi_{ki}} P_i^{l+1}(m,n,r,t) \times \nabla_y \Psi_{ki}^{l+1}(m,n,r,t), \text{ then} \quad (28)$$

$$\Delta y_k^l = \sum_{i=1}^{N_{l+1}} Conv2Dvar\{\Delta_i^{l+1}, \nabla_y P_i^{l+1}(m,n,r,t)\}$$

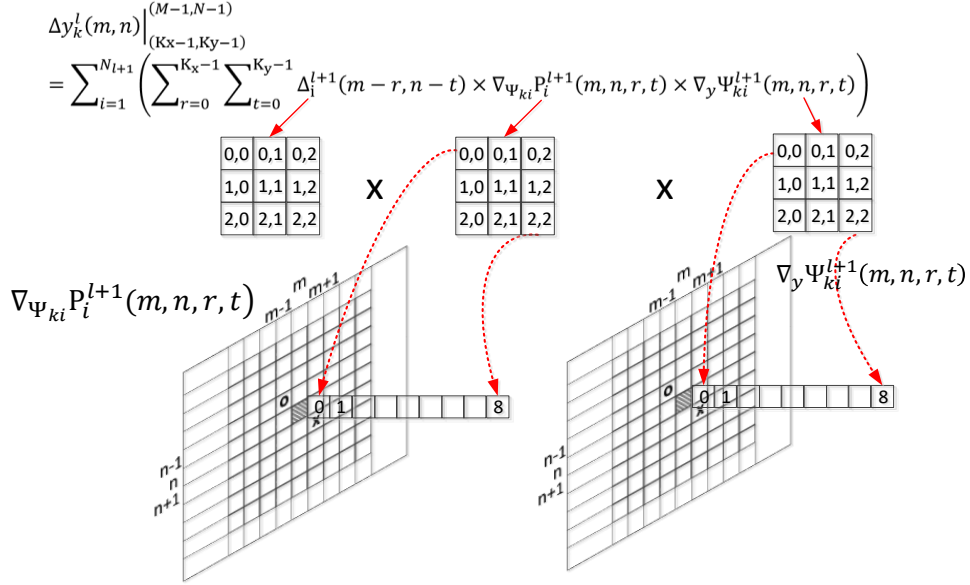

**Figure 7: BP of the delta error from the next operational layer using spatially varying 2D convolutions.**

The two 4-D matrices, $\nabla_{\Psi_{ki}} P_i^{l+1}(m,n,r,t)$ and $\nabla_y \Psi_{ki}^{l+1}(m,n,r,t)$ are illustrated in Figure 7 as the variable (2D) matrices of indices $r$ and $t$, each of which is located at the entry of the 2D matrix of indices, $m$ and $n$. In other words, both $\nabla_{\Psi_{ki}} P_i^{l+1}(m,n,r,t)$ and $\nabla_y \Psi_{ki}^{l+1}(m,n,r,t)$ can be regarded as "varying" kernels with respect to the location, $(m,n)$. Their element-wise multiplication, $\nabla_y P_i^{l+1}(m,n,r,t)$, is also another 4D matrix or equivalently, a varying weight kernel. A closer look to Eq. (28) will reveal the fact that the BP of the delta-errors still yields a "convolution-like" formula as in Eq. (24); however, this time the kernel is not static as in the BP for CNNs, rather a spatially varying kernel with respect to the location $(m,n)$ and hence we call it as the "varying 2D convolution", $Conv2Dvar\{\Delta_i^{l+1}, \nabla_y P_i^{l+1}(m,n,r,t)\}$, of the delta-error and the final varying kernel, $\nabla_y P_i^{l+1}(m,n,r,t)$. As mentioned earlier, when the pool and nodal operators of the linear convolution are used, then: $\nabla_{\Psi_{ki}} P_i^{l+1}(m,n,r,t) = 1$, $\nabla_y P_i^{l+1}(m,n,r,t) = \nabla_y \Psi_{ki}^{l+1}(m,n,r,t) = w_{ik}^{l+1}(r,t)$, which is no longer a varying kernel; hence Eq. (28) will be identical to Eq. (24).

3) *Intra-BP in an ONN neuron:* $\Delta_k^l \xleftarrow{BP} \Delta y_k^l$

This phase is also common for CNNs and ONNs. If there is no up- or down-sampling (pooling) performed within the neuron, once the delta-errors are back-propagated from the next layer, $l+1$, to the neuron in the current layer, $l$, then we can further back-propagate it to the input delta. One can write:

$$\Delta_k^l = \frac{\partial E}{\partial x_k^l} = \frac{\partial E}{\partial y_k^l} \frac{\partial y_k^l}{\partial x_k^l} = \frac{\partial E}{\partial y_k^l} f'(x_k^l) = \Delta y_k^l f'(x_k^l) \quad (29)$$

where $\Delta y_k^l$ is computed as in Eq. (28). On the other hand, if there is a down-sampling by factors, *ssx* and *ssy*, then the back-propagated delta-error by Eq. (28) should be first up-sampled to compute the delta-error of the neuron. Let zero order up-sampled map be: $uy_k^l = \underset{ssx,ssy}{\text{up}}(y_k^l)$. Then Eq. (29) can be updated as follows:

$$\Delta_k^l = \frac{\partial E}{\partial x_k^l} = \frac{\partial E}{\partial y_k^l} \frac{\partial y_k^l}{\partial x_k^l} = \frac{\partial E}{\partial y_k^l} \frac{\partial y_k^l}{\partial uy_k^l} \frac{\partial uy_k^l}{\partial x_k^l}$$
$$= \underset{ssx,ssy}{\text{up}}(\Delta y_k^l) \beta f'(x_k^l) \quad (30)$$

where $\beta = \frac{1}{ssx.ssy}$ since each pixel of $y_k^l$ is now obtained by averaging (*ssx.ssy*) number of pixels of the intermediate output, $uy_k^l$. Finally, if there is a up-sampling by factors, *usx* and *usy*, then let the average-pooled map be: $dy_k^l = \underset{usx,usy}{\text{down}}(y_k^l)$. Then Eq. (29) can be updated as follows:

$$\Delta_k^l = \frac{\partial E}{\partial x_k^l} = \frac{\partial E}{\partial y_k^l} \frac{\partial y_k^l}{\partial x_k^l} = \frac{\partial E}{\partial y_k^l} \frac{\partial y_k^l}{\partial dy_k^l} \frac{\partial dy_k^l}{\partial x_k^l}$$
$$= \underset{usx,usy}{\text{down}}(\Delta y_k^l) \beta^{-1} f'(x_k^l) \quad (31)$$

*4) Computation of the Weight (Kernel) and Bias Sensitivities*

The first three BP stages are performed to compute and back-propagate the delta errors, $\Delta_k^l = \frac{\partial E}{\partial x_k^l}$, to each operational neuron at each hidden layer. As illustrated in Figure 3, a delta error is a 2D map whose size is identical to the input map of the neuron. The sole purpose of back-propagating the delta-errors at each BP iteration is to use them to compute the weight and bias sensitivities. This is evident in the regular BP on MLPs, i.e.:

$$x_i^{l+1} = b_i^{l+1} + .. + y_k^l w_{ik}^{l+1} + ..$$
$$\therefore \frac{\partial E}{\partial w_{ik}^{l+1}} = y_k^l \, \Delta_i^{l+1} \quad and \quad \frac{\partial E}{\partial b_i^{l+1}} = \Delta_i^{l+1} \tag{32}$$

Eq. (14) shows the direct role of the delta-errors in computing the weight and bias sensitivities in a GOP network. To extend this first for the convolutional neurons in a CNN, and then for the operational neurons of an ONN we can follow a similar approach. Figure 8 illustrates the convolution of the output of the current layer neuron, $y_k^l$, and kernel, $w_{ki}^l$, to form the input of the $i^{th}$ neuron, $x_i^{l+1}$, at the next layer, $l+1$.

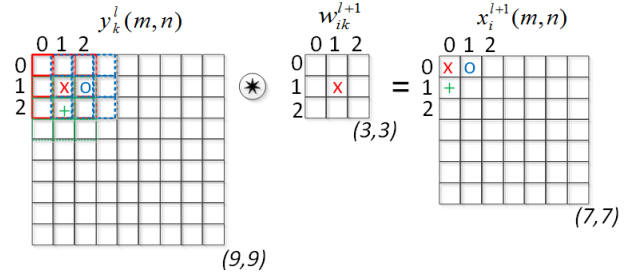

**Figure 8: Convolution of the output of the current layer neuron, $y_k^l$, and kernel, $w_{ki}^l$, to form the input of the $i^{th}$ neuron, $x_i^{l+1}$, at the next layer, $l+1$.**

We can express the contribution of each kernel element over the output as shown in Eq. (33). Since each weight (kernel) element is used in common to form each neuron input, $x_i^{l+1}(m,n)$, the derivative will be the accumulation of delta-output product for all pixels as expressed in Eq. (34). A closer look to the equation will reveal the fact that this dot-product accumulation is actually nothing but a 2D convolution of the output with the delta-error map of the input. It is interesting to notice the parallel relation between the primary transformation and the weight sensitivity computation, i.e., in MLPs it is a scalar multiplication of the weight and output, which is repeated in Eq. (32) between delta-error and output, and now the linear convolution repeats this relation in Eq. (34).

$$\begin{aligned}
x_i^{l+1}(0,0) &= w_{ik}^{l+1}(0,0)y_k^l(0,0) + w_{ik}^{l+1}(0,1)y_k^l(0,1) + w_{ik}^{l+1}(1,0)y_k^l(1,0) + .. \\
x_i^{l+1}(0,1) &= w_{ik}^{l+1}(0,0)y_k^l(0,1) + w_{ik}^{l+1}(0,1)y_k^l(0,2) + w_{ik}^{l+1}(1,0)y_k^l(1,1) + .. \\
x_i^{l+1}(1,0) &= w_{ik}^{l+1}(0,0)y_k^l(1,0) + w_{ik}^{l+1}(0,1)y_k^l(1,1) + w_{ik}^{l+1}(1,0)y_k^l(2,0) + .. \\
&\ldots \\
x_i^{l+1}(m-r,n-t) &= w_{ik}^{l+1}(0,0)y_k^l(m-r,n-t) + \cdots + w_{ik}^{l+1}(r,t)y_k^l(m,n) + .. \\
\therefore x_i^{l+1}(m,n)\Big|_{(0,0)}^{(M-2,N-2)} &= \sum_{r=0}^{2}\sum_{t=0}^{2} w_{ik}^{l+1}(r,t)\, y_k^l(m+r,n+t) + ..
\end{aligned} \tag{33}$$

$$\frac{\partial E}{\partial w_{ik}^{l+1}(r,t)}\Big|_{(0,0)}^{(2,2)} = \sum_{m=r}^{M+r-K_x}\sum_{n=t}^{N+t-K_y} \frac{\partial E}{\partial x_i^{l+1}(m-r,n-t)} \frac{\partial x_i^{l+1}(m-r,n-t)}{\partial w_{ik}^{l+1}(r,t)} = \sum_{m=0}^{M-1}\sum_{n=0}^{N-1} \Delta_i^{l+1}(m-r,n-t)\, y_k^l(m,n)$$
$$\Rightarrow \frac{\partial E}{\partial w_{ik}^{l+1}} = conv2D(y_k^l, \Delta_i^{l+1}, 'NoZeroPad') \tag{34}$$

Finally, the bias for this neuron, $b_k^l$, contributes to all pixels in the image (same bias shared among all pixels), so its sensitivity will be the accumulation of individual pixel sensitivities as expressed in Eq. (35):

$$\frac{\partial E}{\partial b_k^l} = \sum_{m=0}^{M-1}\sum_{n=0}^{N-1} \frac{\partial E}{\partial x_k^l(m,n)} \frac{\partial x_k^l(m,n)}{\partial b_k^l}$$
$$= \sum_{m=0}^{M-1}\sum_{n=0}^{N-1} \Delta_k^l(m,n) \tag{35}$$

Eq. (36) shows the contribution of bias and weights to the next level input map. $x_i^{l+1}(m,n)$. Since bias contribution is a scalar addition, same for the CNN's, the bias sensitivity expression in Eq. (35) can be used for ONNs too. In order to derive the expression for the weight sensitivities we can follow the same approach as before: since each kernel element, $w_{ik}^{l+1}(r,t)$ contributes *all* the pixels of the input pixels, $x_i^{l+1}(m,n)$, by using the chain rule, the weight sensitivities can first be expressed as in Eq. (37) and then simplified into the final form in Eq. (38).

$$Recall: x_i^{l+1}(m-r,n-t)\Big|_{(K_x,K_y)}^{(M-1,N-1)} = b_i^{l+1} + \sum_{k=1}^{N_1} P_i^{l+1}\big[\ldots, \Psi_i^{l+1}\big(w_{ik}^{l+1}(r,t), y_k^l(m,n)\big), \ldots\big] \tag{36}$$

$$\therefore \frac{\partial E}{\partial w_{ik}^{l+1}}(r,t)\Big|_{(0,0)}^{(Kx-1,Ky-1)} =$$

$$\sum_{m=r}^{M+r-Kx} \sum_{n=t}^{N+t-Ky} \left( \begin{array}{c} \frac{\partial E}{\partial x_1^{l+1}(m-r,n-r)} \times \frac{\partial x_1^{l+1}(m-r,n-t)}{\partial P_i^{l+1}[\Psi_i^{l+1}(y_k^l(m-r,n-t),w_{ik}^{l+1}(0,0)),..,\Psi_i^{l+1}(y_k^l(m,n),w_{ik}^{l+1}(r,t),)...]} \times \\ \frac{\partial P_i^{l+1}\left[\Psi_i^{l+1}\left(y_k^l(m-r,n-t),w_{ik}^{l+1}(0,0)\right),...,\Psi_i^{l+1}(y_k^l(m,n),w_{ik}^{l+1}(r,t),)...\right]}{\partial \Psi_{ik}^{l+1}\left(y_k^l(m,n),w_{ik}^{l+1}(r,t)\right)} \times \\ \frac{\partial \Psi_{ik}^{l+1}(y_k^l(m,n),w_{ik}^{l+1}(r,t))}{\partial w_{ik}^{l+1}(r,t)} \end{array} \right) \quad (37)$$

where $\frac{\partial x_1^{l+1}(m-r,n-t)}{\partial P_i^{l+1}[\Psi_i^{l+1}(y_k^l(m-r,n-t),w_{ik}^{l+1}(0,0)),...,\Psi_i^{l+1}(y_k^l(m,n),w_{ik}^{l+1}(r,t),)...]} = 1$.

Let $\nabla_w \Psi_{ki}^{l+1}(m,n,r,t) = \frac{\partial \Psi_{ik}^{l+1}\left(y_k^l(m,n),w_{ik}^{l+1}(r,t)\right)}{\partial w_{ik}^{l+1}(r,t)}$, then it simplifies to:

$$\frac{\partial E}{\partial w_{ik}^{l+1}}(r,t)\Big|_{(0,0)}^{(Kx-1,Ky-1)} = \sum_{m_0=r}^{M+r-Kx} \sum_{n_0=t}^{N+t-Ky} \Delta_1^{l+1}(m_0-r,n_0-t) \times \nabla_{\Psi_{ki}} P_i^{l+1}(m_0,n_0,r,t) \times \nabla_w \Psi_{ki}^{l+1}(m_0,n_0,r,t)$$

Let $\nabla_w P_i^{l+1}(m_0,n_0,r,t) = \nabla_{\Psi_{ki}} P_i^{l+1}(m_0,n_0,r,t) \times \nabla_w \Psi_{ki}^{l+1}(m_0,n_0,r,t)$,

$$\frac{\partial E}{\partial w_{ik}^{l+1}}(r,t)\Big|_{(0,0)}^{(Kx-1,Ky-1)} = \sum_{m_0=r}^{M+r-Kx} \sum_{n_0=t}^{N+t-Ky} \Delta_1^{l+1}(m_0-r,m_0-t) \times \nabla_w P_i^{l+1}(m_0,n_0,r,t), \quad \text{or let } m = m_0 - r, n = n_0 - \quad (38)$$

$$\frac{\partial E}{\partial w_{ik}^{l+1}}(r,t)\Big|_{(0,0)}^{(Kx-1,Ky-1)} = \sum_{m=0}^{M-Kx} \sum_{n=0}^{N-Ky} \Delta_1^{l+1}(m,n) \times \nabla_w P_i^{l+1}(m+r,n+t,r,t)$$

$$\therefore \frac{\partial E}{\partial w_{ik}^{l+1}} = Conv2Dvar(\Delta_i^{l+1}, \nabla_w P_i^{l+1})$$

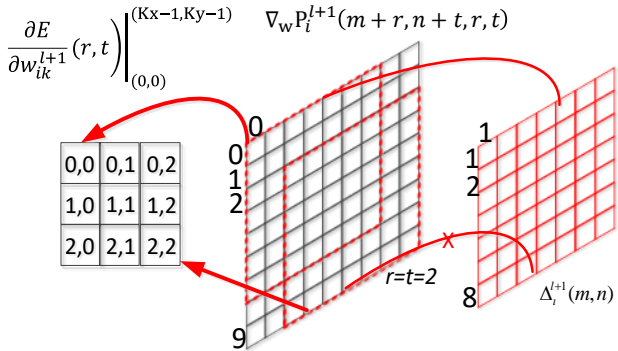

**Figure 9: Computation of the kernel sensitivities.**

Note that the first term, $\Delta_1^{l+1}(m,n)$, in Eq. (38) is a 2D map (matrix) independent from the kernel indices, $r$ and $t$. It will be element-wise multiplied by the other two latter terms, each with the same dimension, (i.e., M-2xN-2 for Kx=Ky=3) and created by derivative functions of nodal and pool operators applied over the pixels of the MxN output, $y_k^l(m,n)$, and the corresponding weight value, $w_{ik}^{l+1}(r,t)$. Note that for each shift value, $r$ and $t$, the weight is fixed, $w_{ik}^{l+1}(r,t)$; however, the pixels are taken from different (shifted) sections of $y_k^l(m,n)$. This operation is illustrated in Figure 9. Finally, it is easy to see that when the pool and nodal operators of convolutional neurons are used, $\nabla_{\Psi_{ki}} P_i^{l+1}(m,n,r,t) = 1$, $\nabla_w \Psi_{ki}^{l+1}(m,n,r,t) = y_k^l(m,n)$, and thus Eq. (38) simplifies to Eq. (34).

### B. Implementation

To bring an ONN to a run-time functionality both FP and BP operations should properly be implemented based on the four phases detailed earlier. Then the optimal operator set per neuron in the network can be searched by short BP training sessions with potential operator set assignments. Finally, the ONN with the best operators can be trained over the train dataset of the problem.

As a typical stochastic gradient descent method, BP has an iterative process where at each iteration, first a forward-propagation (FP) is performed by using the latest kernel weights and biases that are updated during the last BP iteration. During the FP, the required derivatives and sensitivities for BP such as $f'(x_k^l)$, $\nabla_y \Psi_{ki}^{l+1}(m,n,r,t)$, $\nabla_{\Psi_{ki}} P_i^{l+1}(m,n,r,t)$ and $\nabla_w \Psi_{ki}^{l+1}(m,n,r,t)$ need to be computed and stored. In order to accomplish this, we form a temporary 4D matrix, $\Psi x_i^{l+1}(m,n,r,t) = \Psi_i^{l+1}(y_k^l(m+r,n+t),w_{ik}^{l+1}(r,t))$, $\forall r,t \in [0,2]$ and $\forall m,n \in (M-2,N-2)$ for each pixel of the input map, $x_i^{l+1}(m,n)$. It will then be used in the pool operator to create the input map, $x_i^{l+1}(m,n)$, at the next layer. Basically, the pool operator will create a 2D matrix out of a 4D matrix. This is illustrated in Figure 10.

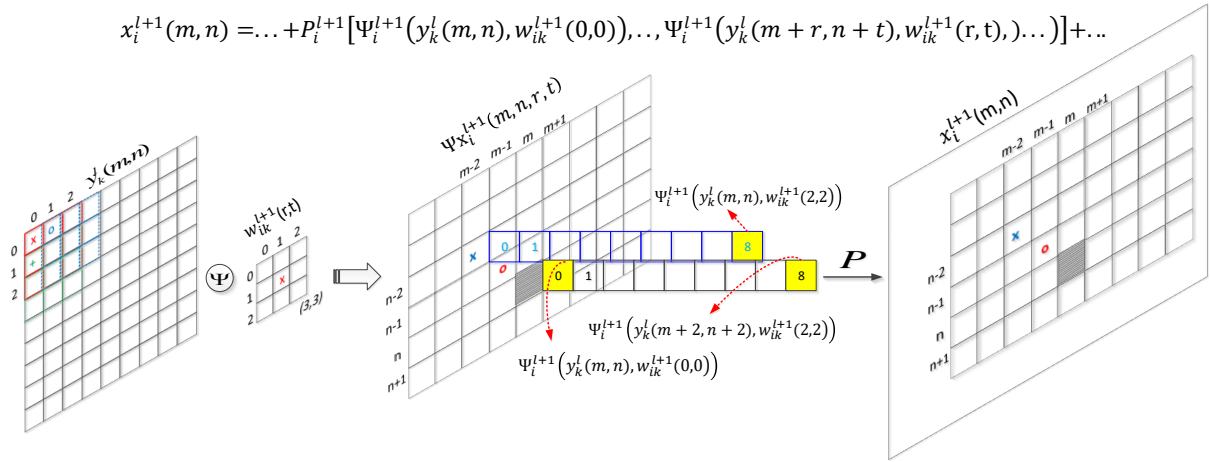

**Figure 10:** The formation of $\Psi x_i^{l+1}(m,n,r,t)$, and the computation of $x_i^{l+1}(m,n)$.

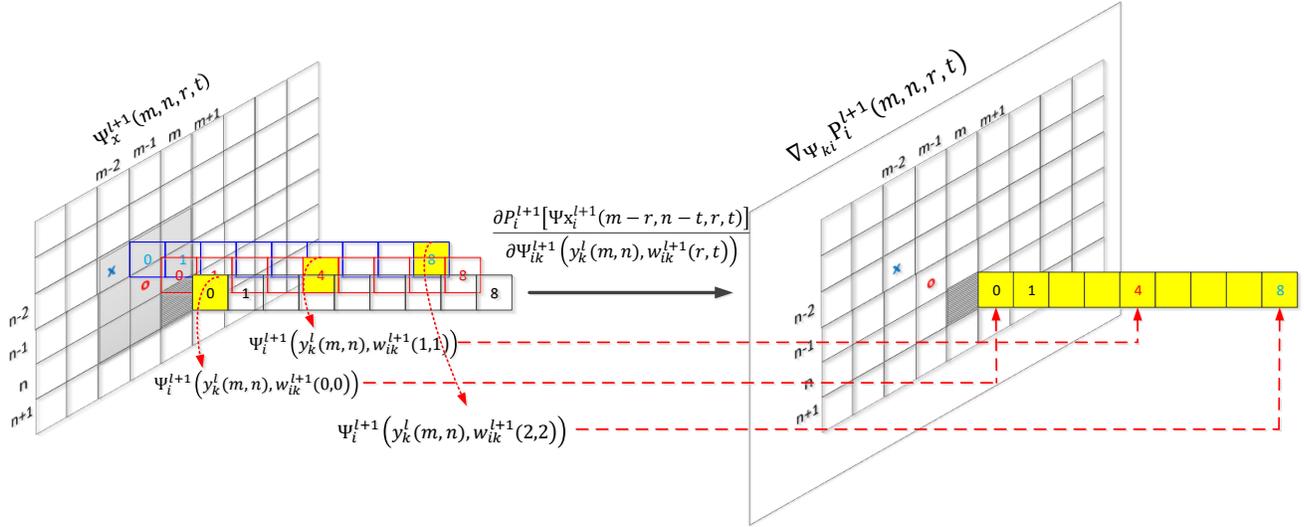

**Figure 11:** Computation of the 4D derivative, $\nabla_{\Psi_{ki}} P_i^{l+1}(m,n,r,t)$, map out of $\Psi x_i^{l+1}(m,n,r,t)$

Once the $\Psi x_i^{l+1}(m,n,r,t)$ is formed, then the 4D derivative $\nabla_{\Psi_{ki}} P_i^{l+1}(m,n,r,t) = \frac{\partial P_i^{l+1}[\Psi x_i^{l+1}(m-r,n-t,r,t)]}{\partial \Psi_{ik}^{l+1}(y_k^l(m,n),w_{ik}^{l+1}(r,t))}$, can easily be composed by computing the derivative of the terms, $\Psi_{ik}^{l+1}(y_k^l(m,n), w_{ik}^{l+1}(r,t))$, which is the only term that contains, $y_k^l(m,n)$, in each $x_i^{l+1}(m,n)$ pixel computation out of pooling the $\Psi x_i^{l+1}(m,n,r,t)$. Figure 11 illustrates this composition. Note that if the pool operator is "summation", then the pool operator derivative, $\nabla_{\Psi_{ki}} P_i^{l+1}(m,n,r,t) = 1$ and hence there is no need to compute and store them. If the nodal operator is "multiplication", then the two 4D nodal derivatives simplify to 2D (static) maps, i.e., $\nabla_y \Psi_{ki}^{l+1}(m,n,r,t) = w_{ik}^{l+1}(r,t)$, and $\nabla_w \Psi_{ki}^{l+1}(m+r,n+t,r,t) = y_k^l(m+r,n+t)$. For any other pool and nodal operator settings, all four 4D maps should be computed during the FP before each BP iteration.

For instance, assume that for an arbitrary operational neuron the pool operator is *Median* and nodal operator is a *Sinosoid*, i.e., $\Psi_{ik}^{l+1}(y_k^l(m,n), w_{ik}^{l+1}(r,t)) = \sin(K. y_k^l(m,n). w_{ik}^{l+1}(r,t))$ where $K$ is a constant. Then the two nodal derivatives will be:

$\nabla_y \Psi_{ki}^{l+1}(m,n,r,t) = K w_{ik}^{l+1}(r,t) \cos(K. y_k^l(m,n). w_{ik}^{l+1}(r,t))$,
$\nabla_w \Psi_{ki}^{l+1}(m+r,n+t,r,t) =$
$\quad K y_k^l(m+r,n+t) \cos(K. y_k^l(m+r,n+t). w_{ik}^{l+1}(r,t))$

Therefore, both 4D derivative matrices are computed first during the FP using the latest weight, $w_{ik}^{l+1}(r,t)$, and output $y_k^l(m,n)$ and stored in this neuron. Similarly for the pool derivative, $\nabla_{\Psi_{ki}} P_i^{l+1}(m,n,r,t)$, first note that the direct derivative of the $Median(\Psi x_i^{l+1}(m,n,r,t))$ with respect to the term, $\Psi_{ik}^{l+1}(y_k^l(m,n), w_{ik}^{l+1}(r,t))$, will give a 4D map where at each pixel location, $(m,n)$, the 2D map at that location will have all the entries 0, except the one which is the median of them. However, when the 4D derivative map, $\nabla_{\Psi_{ki}} P_i^{l+1}(m,n,r,t)$, is composed by collecting the derivatives that include the term, $\Psi_{ik}^{l+1}(y_k^l(m,n), w_{ik}^{l+1}(r,t))$, at a given pixel location, $(m,n)$, the 2D map at that location may have any number of 0s and 1s since this is the term which is obtained from individual $y_k^l(m,n)$ term derivatives.

The conventional BP iterations are executed iteratively to update the weights (the kernel parameters) and biases of each neuron in the ONN until a stopping criterion has been met such as maximum number of iterations (*iterMax*) or the target classification performance (*CP\**) such as mean-square-error (MSE), classification error (CE) or F1. With a proper learning factor, ε, for each BP iteration, *t*, the update for weight kernel and bias at each neuron, *i*, at layer, *l*, can be expressed as follows:

$$w_{ik}^l(t+1) = w_{ik}^l(t) - \varepsilon \frac{\partial E}{\partial w_{ik}^l}$$
$$b_i^l(t+1) = b_i^l(t) - \varepsilon \frac{\partial E}{\partial b_i^l} \quad (39)$$

As a result, the pseudo-code for BP can be presented as in Algorithm 1.

**Algorithm 1: Back-Propagation algorithm for ONNs**

**Input**: $ONN$, $Stopping\ Criteria\ (iterMax, CP^*)$
**Output**: BP trained $ONN^* = BP(ONN, iterMax, CP^*)$

1) Initialize all parameters (i.e., randomly ~U(-a, a) where a=0.1)
2) UNTIL a stopping criterion is reached, ITERATE:
   a. For each item (or a group of items or all items) in the train dataset, DO:
      i. **FP**: Forward propagate from the input layer to the output layer to find outputs, $y_k^l$ the required derivatives and sensitivities for BP such as $f'(x_k^l)$, $\nabla_y \Psi_{ki}^{l+1}$, $\nabla_{\Psi_{ki}} P_i^{l+1}$ and $\nabla_w \Psi_{ki}^{l+1}$ of each neuron,*k*, at each layer, *l*.
      ii. **BP**: Using Eq. (18) compute delta error at the output layer and then using Eqs. (28) and (29) back-propagate the error back to the first hidden layer to compute delta errors of each neuron, *k*, $\Delta_k^l$ at each layer, *l*.
      iii. **PP**: Find the bias and weight sensitivities using Eqs. (35) and (38), respectively.
      iv. **Update**: Update the weights and biases with the (cumulation of) sensitivities found in previous step scaled with the learning factor, ε, as in Eq. (39):
3) Return $ONN^*$

The final task to form the ONN for the learning problem at hand is the search for the best possible operator set for the neurons of ONN. For this purpose, in this study we adopted the greedy iterative search (GIS) [32] [33] due to its simplicity. Since GIS performs layer-wise pruned search the resultant ONN will have homogenous layers each of which has neurons with a common operator set. Let $\{\theta_N^*\}$, be the operator set library consisting of *N* operator sets where each set has a unique nodal, pool and activation operator. With a given learning objective criterion each pass of GIS seeks for the best operator set for a particular layer while keeping the sets of the other layers intact. To accomplish this, one or few (e.g., $N_{BP} = 2$) short BP runs each with random parameter initialization can be performed with each operator set assigned to that level. The operator set, which yields ONN to achieve best performance is then assigned to that layer and the GIS continues with the next layer and the pass continues until the search is carried out for all layers. While always keeping the best operator set assigned to each layer, few GIS passes will suffice to form a near optimal ONN network, $ONN^*(\theta)$, which can then be trained by BP only if the learning objective has not yet been accomplished during the GIS passes. Otherwise, the GIS stops abruptly whenever the learning objective for the problem in hand is met. The pseudo-code for a two-pass GIS is given in Alg. 1.

**Algorithm 2: Two-pass GIS**

**Input**: $\{\theta_N^*\}$, $Stopping\ Criteria\ (iterMax, CP^*)$, $N_{BP}$
**Output**: $ONN^*(\theta)$

1) **Initialization**: Form an ONN with neurons having operator set (nodal, pool and activation) randomly selected from $\{\theta_N^*\}$:
2) For GIS-pass = 1:2 DO:
   a. Starting from output layer, *l*=L:1, DO:
      i. For $\forall \theta_i \in \{\theta_N^*\}$ DO:
         1. **Assign** the operator set of each neuron in the *l*th layer of the ONN to $\theta_i$ → ONN(*l*, $\theta_i$)
         2. **Perform**: $N_{BP}$ × BP(ONN(*l*, $\theta_i$), *iterMax*, *CP\**) and **Record**: ONN*($\theta_i$) that achieves the best performance
      ii. **Assign** $\theta_i^*$ as the operator set of each neuron in the *l*th layer of the ONN → ONN(*l*, $\theta_i^*$)
      iii. **Check**: If *CP\** is reached in any BP run, **break** GIS
3) **RETURN**: ONN*($\theta$) the best performing ONN.

IV. EXPERIMENTAL RESULTS

In this section we perform comparative evaluations between conventional CNNs and ONNs over four challenging problems: 1) Image Syntheses, 2) Denoising, 3) Face Segmentation, and 4) Image Transformation. In order to demonstrate the learning capabilities of the ONNs better, we have further taken the following restrictions:

i) Low Resolution: We keep the image resolution very low, e.g., thumbnail size (i.e., 60x60 pixels) which makes especially pattern recognition tasks (e.g. face segmentation) even harder.
ii) Compact Model: We keep the ONN configuration compact, e.g., only two hidden layers with less than 50 hidden neurons, i.e., *Inx16x32xOut*. Moreover, we shall keep the output layer as a convolutional layer whilst optimizing only the two hidden layers by GIS.
iii) Scarce Train Data: For the two problems (image denoising and segmentation) with train and test datasets, we shall train the network over a limited data (i.e., only 10% of the dataset) while testing over the rest with a 10-fold cross validation.
iv) Multiple Regressions: For the two regression problems (image syntheses and transformation), a single network will be trained to regress multiple (e.g., 4-8) images.
v) Shallow Training: Maximum number of iterations (*iterMax*) for BP training will be kept low (e.g. max. 80 and 240 iterations for GIS and regular BP sessions, respectively).

For a fair evaluation, we shall first apply the same restrictions over the CNNs; however, we shall then relax them to find out whether CNNs can achieve the same learning performance level with, e.g., more complex configuration with deeper training over the simplified problem.

*A. Experimental Setup*

In any BP training session, for each iteration, *t*, with the MSE obtained at the output layer, *E(t)*, a global adaptation of the learning rate, ε, is performed within the range [5.10⁻¹, 5.10⁻⁵], as follows:

$$\varepsilon(t) = \begin{cases} \alpha\varepsilon(t-1) & \text{if } E(t) < E(t-1) \text{ and } \alpha\varepsilon(t-1) \leq 5 \cdot 10^{-1} \\ \beta\varepsilon(t-1) & \text{if } E(t) \geq E(t-1) \text{ and } \beta\varepsilon(t-1) \geq 5 \cdot 10^{-5} \\ \varepsilon(t-1) & \text{else} \end{cases} \quad (40)$$

where $\alpha=1.05$ and $\beta=0.7$, respectively. Since BP training is a stochastic gradient descent method, for each problem we shall perform 10 BP runs, each with random parameter initialization.

The operator set library that is used to form the ONNs to tackle the challenging learning problems in this study is composed of a few essential nodal, pool and activation operators. Table 1 presents the 7 nodal operators along with their derivatives, $\nabla_w \Psi_{ki}^{l+1}$ and $\nabla_y \Psi_{ki}^{l+1}$ with respect to the weight, $w_{ik}^{l+1}$, and the output, $y_k^l$ of the previous layer neuron. Similarly, Table 2 presents the two common pool operators and their derivatives with respect to the nodal term, $\sum_{k=1}^{N_l} \Psi_i^{l+1}(w_{ik}^{l+1}, y_k^l)$.

**Table 1: Nodal operators and derivatives**

| $i$ | Function | $\Psi_i^{l+1}(y_k^l, w_{ik}^{l+1})$ | $\nabla_w \Psi_{ki}^{l+1}$ | $\nabla_y \Psi_{ki}^{l+1}$ |
|---|---|---|---|---|
| 0 | **Mul.** | $w_{ik}^{l+1} y_k^l$ | $y_k^l$ | $w_{ik}^{l+1}$ |
| 1 | **Cubic** | $K w_{ik}^{l+1} (y_k^l)^3$ | $K(y_k^l)^3$ | $3K w_{ik}^{l+1} (y_k^l)^2$ |
| 2 | **Harmonic** | $\sin(K w_{ik}^{l+1} y_k^l)$ | $K y_k^l \cos(K w_{ik}^{l+1} y_k^l)$ | $K w_{ik}^{l+1} \cos(K w_{ik}^{l+1} y_k^l)$ |
| 3 | **Exp.** | $\exp(w_{ik}^{l+1} y_k^l) - 1$ | $y_k^l \exp(w_{ik}^{l+1} y_k^l)$ | $w_{ik}^{l+1} \exp(w_{ik}^{l+1} y_k^l)$ |
| 4 | **DoG** | $w_{ik}^{l+1} y_k^l \exp(-K_D(w_{ik}^{l+1})^2 (y_k^l)^2)$ | $y_k^l \left(1 - 2K_D(w_{ik}^{l+1})^2 (y_k^l)^2\right) \exp(-K_D(w_{ik}^{l+1})^2 (y_k^l)^2)$ | $w_{ki}^l \left(1 - 2K_D(w_{ik}^{l+1})^2 (y_k^l)^2\right) \exp(-K_D(w_{ik}^{l+1})^2 (y_k^l)^2)$ |
| 5 | **Sinc** | $\sin(K w_{ik}^{l+1} y_k^l) / y_k^l$ | $K \cos(K w_{ik}^{l+1} y_k^l)$ | $(K w_{ik}^{l+1} \cos(K w_{ik}^{l+1} y_k^l) / y_k^l) - (\sin(K w_{ik}^{l+1} y_k^l) / (y_k^l)^2)$ |
| 6 | **Chirp** | $\sin(K_C w_{ik}^{l+1} (y_k^l)^2)$ | $K_C (y_k^l)^2 \cos(K w_{ik}^{l+1} (y_k^l)^2)$ | $2K_C w_{ik}^{l+1} y_k^l \cos(K_C w_{ik}^{l+1} (y_k^l)^2)$ |

**Table 2: Pool operators and derivatives**

| $i$ | Function | $P_i^{l+1}[\ldots, \Psi_i^{l+1}(y_k^l, w_{ik}^{l+1}), \ldots]$ | $\nabla_{\Psi_{ki}} P_i^{l+1}$ |
|---|---|---|---|
| 0 | **Summation** | $\sum_{k=1}^{N_l} \Psi_i^{l+1}(w_{ik}^{l+1}, y_k^l)$ | 1 |
| 1 | **Median** | $\underset{k}{\text{median}}(\Psi_i^{l+1}(w_{ik}^{l+1}, y_k^l))$ | $\begin{cases} 1 & \text{if } \arg\text{median}(\Psi_i^{l+1}(w_{ik}^{l+1}, y_k^l)) = k \\ 0 & \text{else} \end{cases}$ |

**Table 3: Activation operators and derivatives**

| $i$ | Function | $f(x)$ | $f'(x)$ |
|---|---|---|---|
| 0 | **Tangent hyperbolic** | $\tanh(x) = \dfrac{1 - e^{-2x}}{1 + e^{-2x}}$ | $1 - f(x)^2$ |
| 1 | **Linear-Cut** | $\text{lin-cut}(x) = \begin{cases} x/cut & \text{if } |x| \leq cut \\ -1 & \text{if } x < -cut \\ 1 & \text{if } x > cut \end{cases}$ | $\begin{cases} 1/cut & \text{if } |x| \leq cut \\ 0 & \text{else} \end{cases}$ |

Finally, Table 3 presents the two common activation functions (operators) and their derivatives. Using these lookup tables, the error at the output layer can be back-propagated and the weight sensitivities can be computed. The top section of Table 4 enumerates each potential operator set and the bottom section presents the index of each individual operator set in the operator library, $\Theta$, which will be used in all experiments. There is a total of $N=7\times 2\times 2=28$ sets that constitute the operator set library, $\{\theta_N^*\}$. Let $\theta_i: \{i_{pool}, i_{act}, i_{nodal}\}$ be the $i^{th}$ operator set in the library. Note that the first operator set, $\theta_0: \{0,0,0\}$ with index $i=0$, belongs to the native operators of a CNN to perform linear convolution with traditional activation function, $tanh$.

In accordance with the activation operators used, the dynamic range of the input/output images in all problems are normalized in the range of [-1, 1] as follows:

$$p_i = 2 \frac{p_i - \min(p)}{\max(p) - \min(p)} - 1 \quad (41)$$

where $p_i$ is the $i^{th}$ pixel value in an image, $p$.

As mentioned earlier, the same compact network configuration with only two hidden layers and a total of 48 hidden neurons, $In\times 16\times 32\times Out$ is used in all the experiments. The first hidden layer applies sub-sampling by $ssx = ssy = 2$, and the second one applies up-sampling by $usx = usy = 2$.

**Table 4: Operator enumeration (top) and the index of each operator set (bottom).**

| i     | 0    | 1       | 2   | 3   | 4   | 5    | 6     |
|-------|------|---------|-----|-----|-----|------|-------|
| Pool  | sum  | median  |     |     |     |      |       |
| Act.  | tanh | lin-cut |     |     |     |      |       |
| Nodal | mul. | cubic   | sin | exp | DoG | sinc | chirp |

| Θ Index | Pool | Act. | Nodal |
|---------|------|------|-------|
| 0       | 0    | 0    | 0     |
| 1       | 0    | 0    | 1     |
| 2       | 0    | 0    | 2     |
| 3       | 0    | 0    | 3     |
| 4       | 0    | 0    | 4     |
| 5       | 0    | 0    | 5     |
| 6       | 0    | 0    | 6     |
| 7       | 0    | 1    | 0     |
| 8       | 0    | 1    | 1     |
| ...     | ...  | ...  | ...   |
| 26      | 1    | 1    | 5     |
| 27      | 1    | 1    | 6     |

*B. Evaluation of the Learning Performance*

In order to evaluate the learning performance of the ONNs for the regression problems, image denoising, syntheses and transformation, we used the Signal-to-Noise Ratio (SNR) evaluation metric, which is defined as the ratio of the signal power to noise power, i.e., $SNR = 10\log(\sigma_{signal}^2/\sigma_{noise}^2)$. The ground-truth image is the original signal and its difference to the actual output yields the "noise" image. For the (face) segmentation problem, with train and test partitions, we used the conventional evaluation metrics such as classification error (*CE*) and *F1*. Given the ground-truth segmentation mask, the final segmentation mask is obtained from the actual output of the network by SoftMax thresholding. With a pixel-wise comparison, *Accuracy* (*Acc*), which is the ratio of the number of correctly classified pixels to the total number of pixels, *Precision* (*P*), which is the rate of correctly classified object (face) pixels in all pixels classified as "face", and *Recall* (*R*), which is the rate of correctly classified "face" pixels among all true "face" pixels can be directly computed. Then $CE = 1 - Acc$ and $F1 = 2PR/(P + R)$. The following sub-sections will now present the results and comparative evaluations of each problem tackled by the proposed ONNs and conventional CNNs.

*1) Image Denoising*

Image denoising is a popular field where deep CNNs have recently been applied and achieved the state-of-the-art performance [36]-[39]. This was an expected outcome since "convolution" is the basis of the linear filtering and a deep CNN with thousands of sub-band filters that can be tuned to suppress the noise in a near-optimal way is a natural tool for image denoising. Therefore, in this particular application we are in fact investigating whether stacked non-linear filters in an ONN can also be tuned for this task and if so, whether it can perform equal or better than its linear counterparts.

In order to perform comparative evaluations, we used the 1500 images from Pascal VOC database. The gray-scaled and down-sampled original images are the target outputs while the images corrupted by and Gaussian White Noise (GWN) are the input. The noise level is kept very high on purpose, i.e., all noisy images have SNR = 0dB. The dataset is then partitioned into train (10%) and test (90%) with 10-fold cross validation. So, for each fold, both network types are trained 10 times by BP over the train (150 images) partition and tested over the rest (1350 images). To evaluate their best learning performances for each fold, we selected the best performing networks (among the 10 BP training runs with random initialization). Then the average performances (over both train and test partitions) of the 10-fold cross validation are compared for the final evaluation.

For ONNs, the layer-wise GIS for best operator set is performed only once (only for the 1st fold) and then the same operator set is used for all the remaining folds. Should it be performed for all the folds, it is likely that different operators sets that could achieve even higher learning performance levels could have been found for ONNs. To further speed up the GIS, as mentioned earlier we keep the output layer as a convolutional layer whilst optimizing only the two hidden layers by GIS. For this problem (over the 1st fold), GIS results in operator indices as 9 for both layers, and it corresponds to the operator indices: 9:{0, 1, 2} for the pool (*summation*=0), activation (*linear-cut*=1) and nodal (*sin*=2), respectively.

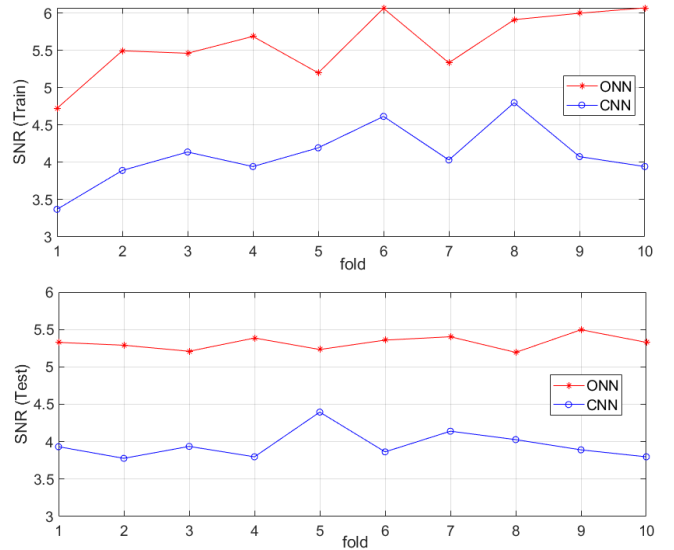

**Figure 12:** Best denoising SNR levels for each fold achieved in train (top) and test (bottom) partitions.

Figure 12 shows SNR plots of the best CNNs and ONNs at each fold over both partitions. Obviously in both train and test partitions ONNs achieved a significant gap around 1.5dB. It is especially interesting to see that although the ONNs are trained over a minority of the dataset (10%), it can still achieve a similar denoising performance in the test set (between 5 to 5.5 dB SNR) while the SNR level of the majority of the (best) CNNs is below 4dB. The average SNR levels of the CNN vs. ONN denoising for the train and test partitions are 5.59dB vs. 4.1dB, and 5.32dB vs. 3.96dB, respectively. For a visual evaluation, Figure 13 shows randomly selected original (target) and noisy (input) images and the corresponding outputs of the best CNNs and ONNs from the test partition. The severe blurring effect of the linear filtering (convolution) is visible at the CNN outputs while ONNs can preserve the major edges despite the severe noise level induced.

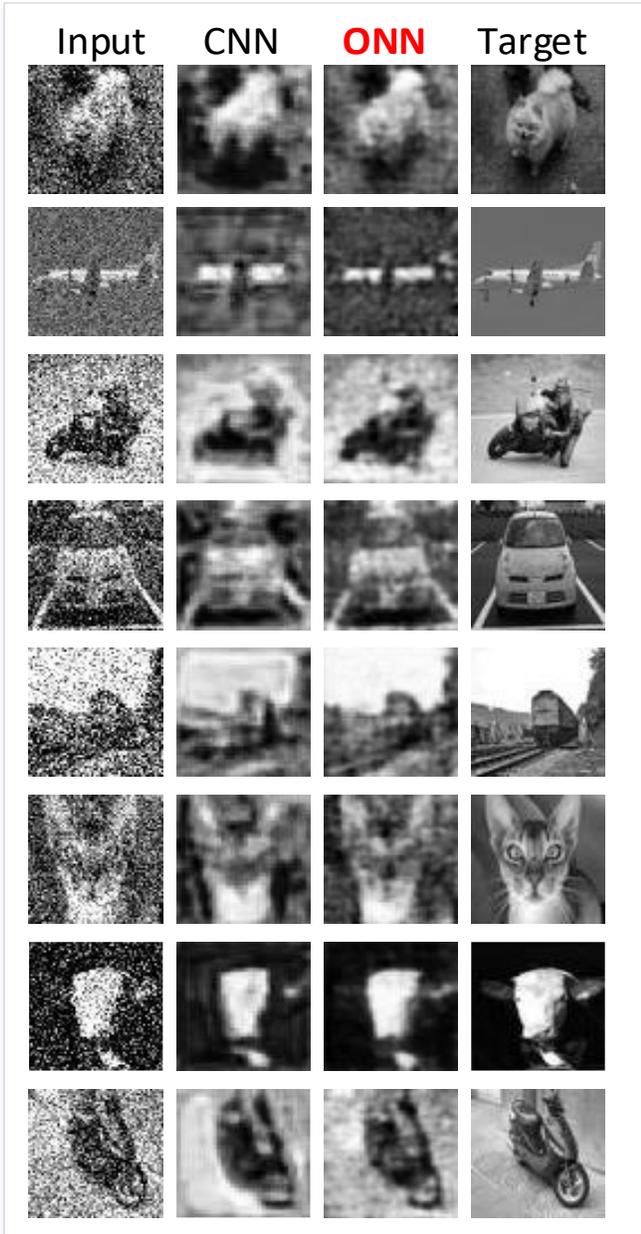

**Figure 13: Some random original (target) and noisy (images) and the corresponding outputs of the best CNN and ONN from the test partition.**

*2) Image Syntheses*

In this problem we aim to test whether a single network can (learn to) synthesize one or many images from WGN images. This is harder than the denoising problem since the idea is to use the noise samples for creating a certain pattern rather than suppressing them. To make the problem even more challenging, we have trained a single network to (learn to) synthesize 8 (target) images from 8 WGN (input) images, as illustrated in Figure 14. We repeat the experiment 10 times (folds), so 8x10=80 images randomly selected from Pascal VOC dataset. The gray-scaled and down-sampled original images are the target outputs while the WGN images are the input. For each trial, we performed 10 BP runs each with random initialization and we select the best performing network for each run for comparative evaluations.

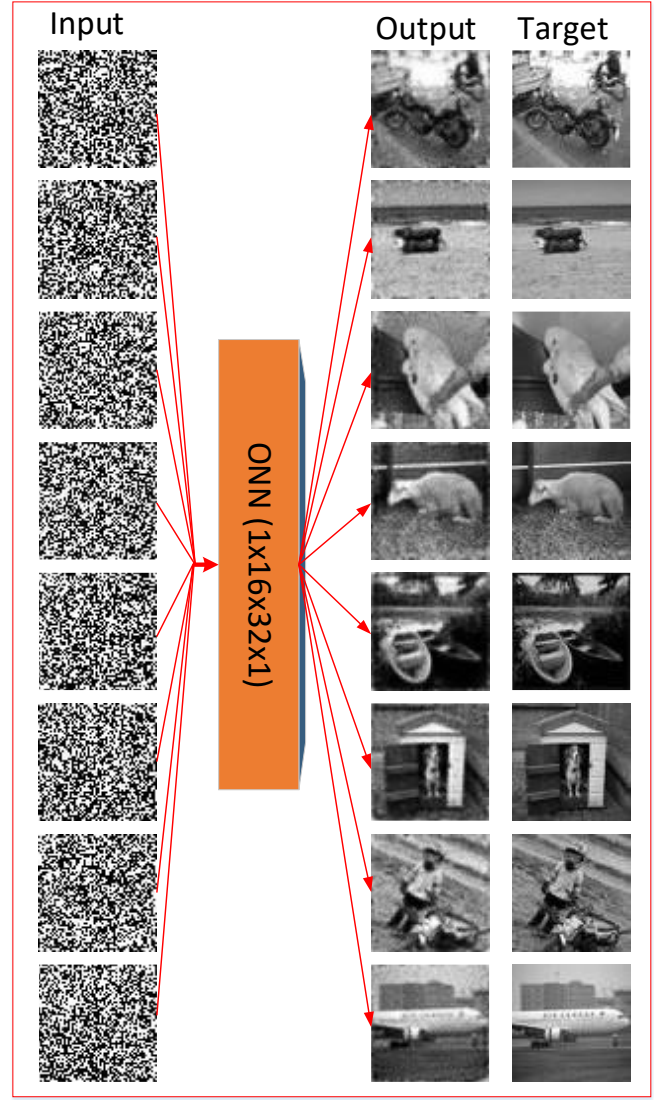

**Figure 14: The outputs of the BP-trained ONN with the corresponding input (WGN) and target (original) images from the 2$^{nd}$ syntheses fold.**

As in the earlier application, the layer-wise GIS for seeking the best operator set is performed only once (only for the 1$^{st}$ fold) for the two hidden operational layers of ONNs and then the same operator set is used for all the remaining folds. Hence over the 1$^{st}$ fold, GIS yields the top-ranked operator set with the operator indices as 3 and 13 for the 1$^{st}$ and 2$^{nd}$ hidden layers, which correspond to the operator indices: 1) (0, 0, 3) for the pool (*summation*=0), activation (*tanh*=0) and nodal (*exp*=3), respectively, and 2) (0, 1, 6) for the pool (*summation*=0), activation (*linear-cut*=1) and nodal (*chirp*=6), respectively.

Figure 15 shows the SNR plots of the best CNNs and ONNs among the 10 BP runs for each syntheses experiment (fold). Several interesting observations can be made from these results. First, the best SNR level that CNNs have ever achieved is below 8dB while this is above 11dB for ONNs. A critical issue is that at the 4$^{th}$ syntheses fold, neither of the BP runs is able to train the CNN to be able to synthesize that batch of 8 images (SNR < -1.6dB). Obviously, it either requires more BP runs than 10 or more likely, it requires a more complex/deeper network configuration. On the other hand, ONNs never failed to achieve a reasonable syntheses performance as the worst SNR level (from fold 3) is still higher than 8dB. The average SNR levels of the CNN and ONN syntheses are 5.02dB and 9.91dB, respectively. Compared to the denoising problem, the performance gap significantly widened since this is now a much harder learning problem. For a visual comparative evaluation, Figure 16 shows a random set of 14 syntheses outputs of the best CNNs and ONNs with the target image. The performance gap is also clear here especially some of the CNN syntheses outputs have suffered from severe blurring and/or textural artefacts.

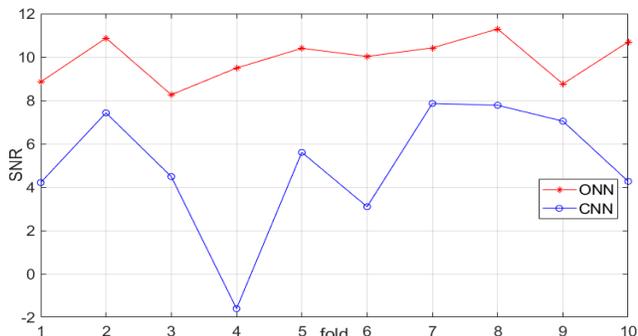

**Figure 15: Best SNR levels for each syntheses fold achieved by CNNs (blue) and ONNs (red).**

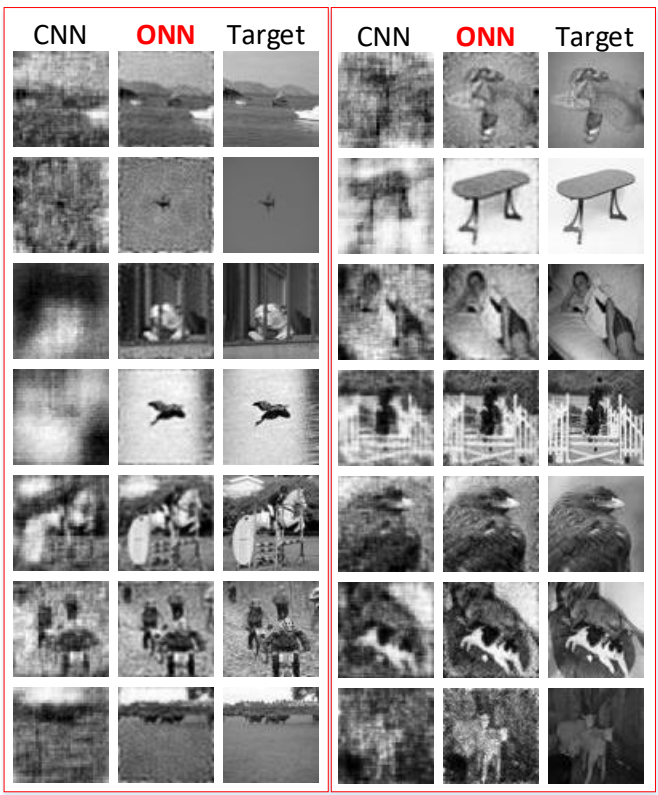

**Figure 16: A random set of 14 syntheses outputs of the best CNNs and ONNs with the target images. The WGN input images are omitted.**

*3) Face Segmentation*

Face or object segmentation (commonly referred as "Semantic Segmentation") in general is a common application domain especially for deep CNNs [41]-[50]. In this case the input is the original image and the output is the segmentation mask which can be obtained by simply thresholding the output of the network. In this section we perform comparative evaluations between CNNs and ONNs for face segmentation. In [41] an ensemble of compact CNNs was tested against a deep CNN and this study has shown that a compact CNN with few convolutional layers and dozens of neurons is capable of learning certain face patterns but may fail for other patterns. This was the reason it was proposed to use an ensemble of compact CNNs in a "Divide and Conquer" paradigm.

In order to perform comparative evaluations, we used FDDB face detection dataset [51]. FDDB dataset contains 1000 images with one or many human faces in each image. We keep the same experimental setup as in image denoising application: The dataset is partitioned them into train (10%) and test (90%) with 10-fold cross validation. So for each fold, both network types are trained 10 times by BP over the train (100 images) partition and tested over the rest (900 images). To evaluate their best learning performances for each fold, we selected the best performing networks (among the 10 BP training runs with random initialization). Then the average performances (over both train and test partitions) of the 10-fold cross validation are compared for the final evaluation.

For ONNs, the layer-wise GIS for best operator set is performed only once (only for the 1$^{st}$ fold). But this time, in order to see the effect of different operator sets on the train and test performance, we selected the top 1$^{st}$ and 3$^{rd}$ ranked operator sets in GIS and used them to create two distinct ONNs. The top ranked operator set has the operator indices as 12 and 2 for the 1$^{st}$ and 2$^{nd}$ hidden layers, which correspond to the operator indices: 1) (0, 1, 5) for the pool (*summation*=0), activation (*lin-cut*=1) and nodal (*sinc*=5), respectively, and 2) (0, 0, 2) for the pool (*summation*=0), activation (*tanh*=0) and nodal (*sin*=2), respectively. The 3$^{rd}$ top ranked operator set has the operator indices as 10 and 9 for the 1$^{st}$ and 2$^{nd}$ hidden layers, which correspond to the operator indices: 1) (0, 1, 3) for the pool (*summation*=0), activation (*lin-cut*=1) and nodal (*exp*=3), respectively, and 2) (0, 1, 2) for the pool (*summation*=0), activation (*lin-cut*=1) and nodal (*sin*=2), respectively. Finally, we label the ONNs with the 1$^{st}$ and 3$^{rd}$ ranked operators' sets as, ONN-1 and ONN-3, respectively.

Figure 17 shows F1 plots of the best CNNs and ONNs at each fold over both partitions. The average F1 scores of the CNN vs. (ONN-1 and ONN-3) segmentation for the train and test partitions are: 58.58% vs. (87.4% and 79.86%), and 56.74% vs. (47,96% and 59.61%), respectively. As expected, ONN-1 has achieved the highest average F1 in all folds on the train partition and this is around 29% higher than the segmentation performance of the CNNs. Despite of its compact configuration, this indicates an "Over-fitting" since its average generalization performance over the test partition is around 8% lower than the average F1 score of CNN. Nevertheless, ONN-3 shows a superior performance level in both train and test partitions by around 21% and 3%, respectively. Since GIS is performed over the train partition, ONN-3 may, too, suffer from over-fitting as there is a significant performance gap between the train and test partitions. This can be addressed, for instance, by performing GIS over a validation set to find out the (near-) optimal operator set that can generalize the best.

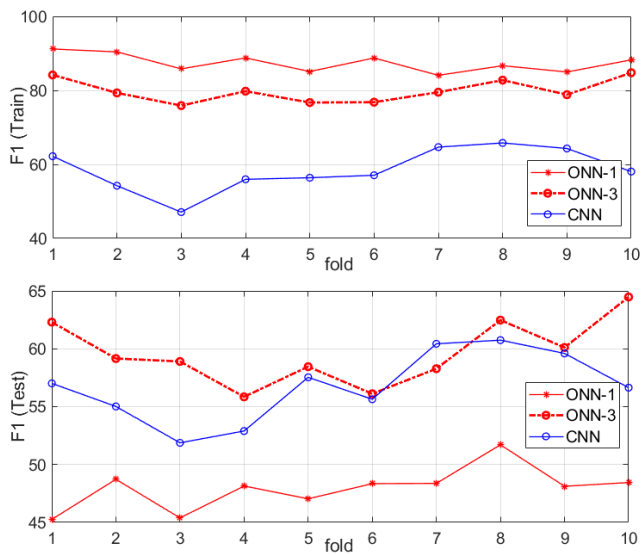

**Figure 17: Best segmentation F1 levels for each fold achieved in train (top) and test (bottom) partitions by ONN-1 (solid-red), ONN-3 (dashed-red) and CNN (solid-blue).**

*4) Image Transformation*

Image transformation (or sometimes called as image translation) is the process of converting one (set of) image(s) to another. Deep CNNs have recently been used for certain image translation tasks [52], [53] such as edge-to-image, gray-scale-to-color image, day-to-night (or vice versa) photo translation, etc. In all these applications, the input and output (target) images are closely related. In this study we tackled a more challenging image transformation, which is transforming an image to entirely different image. This is also much harder than the image syntheses problem because this time the problem is the creation of a (set of) image(s) from another with a distinct pattern and texture.

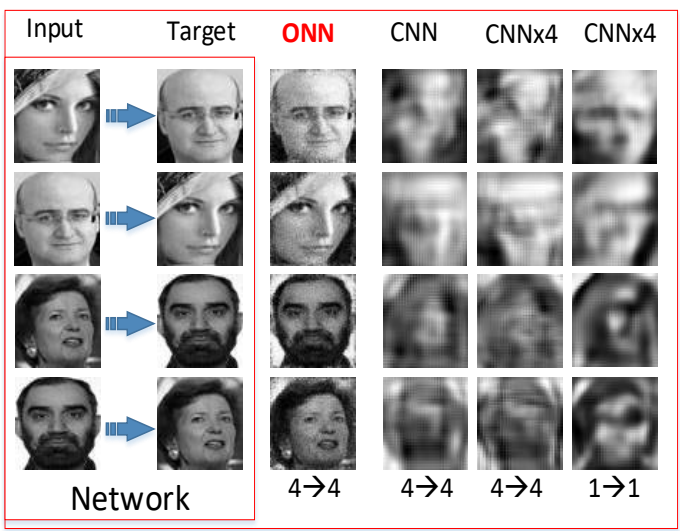

**Figure 18: Image transformation of the 1st fold including two inverse problems (left) and the outputs of the ONN and CNN with the default configuration, and the two CNNs (CNNx4) with 4 times more parameters. On the bottom, the numbers of input → target images are shown.**

To make the problem even more challenging, we have trained a single network to (learn to) transform 4 (target) images from 4 input images, as illustrated in Figure 18 (left). In the first fold, we have further tested whether the networks are capable of learning the "inverse" problems, which means, the same network can transform a pair of input images to another pair of output images and also do the opposite (output images become the input images). Images used in the first fold are shown in Figure 18 (left). We repeat the experiment 10 times using the close-up "face" images most of which obtained from the FDDB face detection dataset [51]. The gray-scaled and down-sampled images are used as both input and output. For CNNs we performed 10 BP runs each with random initialization and for comparative evaluations we select the best performing network for each run. For ONNs, we perform 2-pass GIS for each fold and each BP-run within the GIS is repeated 10 times to evaluate the next operator set assigned.

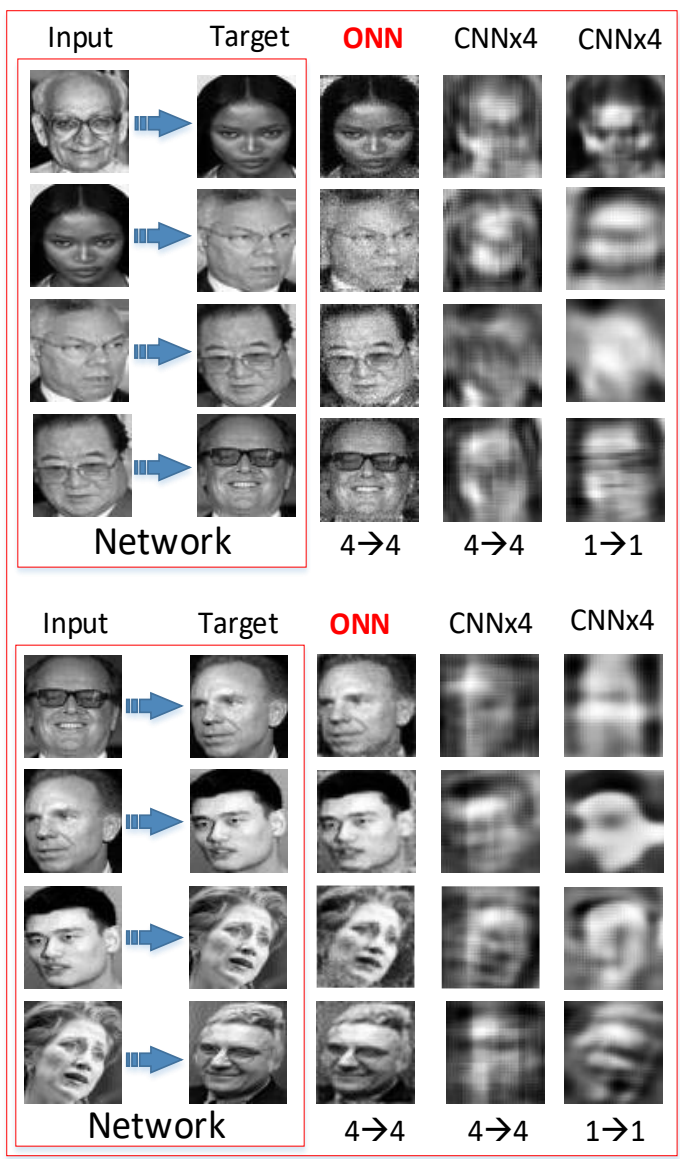

**Figure 19: Image transformations of the 3rd (top) and 4th (bottom) folds and the outputs of the ONN, and the two CNNs (CNNx4) with 4 times more parameters. On the bottom, the numbers of input → target images are shown.**

In the first fold, the outputs of both networks are shown in Figure 18 (right). The GIS results in the optimal operator set that has the operator indices as 0 and 13 for the 1st and 2nd hidden layers, and this corresponds to the operator indices: 1) 0:{0, 0, 0} for the pool (*summation*=0), activation (*tanh*=0) and nodal (*mul*=0), respectively, and 2) 13:{0,1,6} for the pool (*summation*=0), activation (*lin-cut*=1) and nodal (*chirp*=6), respectively. The average SNR level achieved is 10.99 dB, which is one of the highest SNR achieved among all 10 folds despite the fact that in this fold ONNs are trained for the transformation of two *inverse* problems. On the other hand, we had to use three distinct configurations for CNNs. Because the CNN with the default configuration, and the populated configuration, CNNx4 that is a CNN with the number of hidden neurons twice the default number (2x48=96 neurons), both failed to perform a reasonable transformation. Even though CNNx4 has twice as much hidden neurons (i.e., 1x32x64x1) and around 4-times more parameters, the best BP training among 10 runs yield the average SNR = 0.172 dB, which is slightly higher than the average SNR = 0.032 dB obtained by the CNN with the default configuration. Even though we later simplified the problem significantly by training a *single* CNN for transforming only *one* image (rather than 4) whilst still using the CNNx4 configuration, the average SNR improved to 2.45dB which is still far below the acceptable performance level since the output images are still unrecognizable.

Figure 19 shows the results for the image transformations of the 3rd and 4th folds. A noteworthy difference with respect to the 1st fold is that in both folds, the 2-pass GIS results in a different operator set for the 1st hidden layer, which has the operator indices: 1) 3:{0,0,3} for the pool (*summation*=0), activation (*tanh*=0) and nodal (*exp*=3), respectively. The average SNR levels achieved are 10.09 dB and 13.01 dB, respectively. In this figure, we skipped the outputs of the CNN with the default configuration since, as in the 1st fold, it has entirely failed (i.e., average SNRs are -0.19 dB and 0.73 dB, respectively). This is also true for the CNNx4 configuration even though a significant improvement is observed, i.e., average SNR levels are 1.86 dB and 2.37 dB, respectively. An important observation is that, these levels are significantly higher than the corresponding SNR level for the 1st fold since both folds (transformations) are relatively easier than the transformation of the two inverse problems in the 1st fold. However, the transformation quality is still far from satisfactory. Finally, when the problem is significantly simplified as before, that is, a single CNN is trained to learn transformation for only *one* image pair (1→1), then CNNx4 can then achieve the average SNR level of 2.54dB, which still makes it far from being satisfactory. This is true for the remaining folds and over the 10 folds, the average SNR levels for ONNs, CNNs, and the two CNNx4 configurations are: 10.42 dB, -0.083 dB, 0.24 dB (4→4) and 2.77 dB (1→1), respectively. This indicates that a significantly more complex and deeper configuration is needed for CNNs to achieve a reasonable transformation performance.

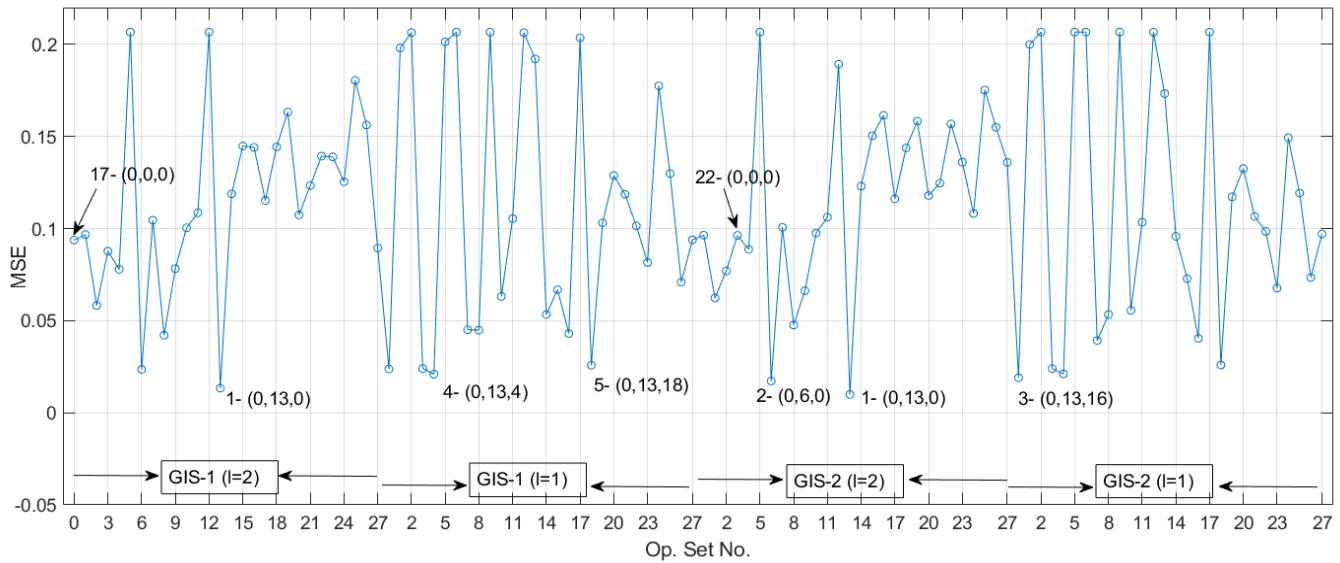

**Figure 20: MSE plot during the 2-pass GIS operation for the 1st fold. The top-5 ranked operator sets found in three layers, (3rd, 2nd, 1st) are shown in parantheses. The native operator set of CNNs, (0, 0, 0), with operator set index 0, can get the 17th and 22nd ranks among the operator sets searched.**

In order to further investigate the role of the operators on the learning performance, we keep the log of operator sets evaluated during the 2-pass GIS. For the 1st fold of the image transformation problem, Figure 20 shows the average MSE obtained during the 2-pass GIS. Note that the output layer's (layer-3) operator set is fixed as, 0:{0,0,0} in advance and excluded from GIS. This plot clearly indicates which operator sets are the best-suited for this problem and which are not. Obviously the operator sets with indices, 6:{0, 0, 6} and 13:{0, 1, 6} in layer-2 got the top ranks, both of which use the pool operator *summation* and the nodal operator, *chirp*. For layer-1, both of them favors the operator set with index 0:{0,0,0}. Interestingly, the 3rd ranked operator set is, 13:{0,1,6} in layer-2, and 16:{1,2,6} in layer-1, for the pool (*median*=1), activation (*tanh*=0) and nodal (*sin*=2), respectively. The pool operator, *median*, is also used in the 5th ranked operator set for layer-1 too. For all the problems tackled in this study, although it never got to the top-ranked operator set for any layer, it has obtained 2nd or 3rd ranks in some of the problems. Finally, an important observation worth mentioning is the ranking of the native operator set of a CNN with operator index, 0:{0,0,0}, which was evaluated twice during the 2-pass GIS. In both evaluations, among the 10 BP runs performed, the minimum MSE obtained was close to 0.1 which makes it the 17th and 22nd best operator set among all the sets evaluated. This means that there are *at least* 16 operator sets (or equivalently 16 distinct ONN models each with different operator sets but same network configuration) which will yield a better transformation performance than the CNN's. This is, in fact, a "best-case" scenario for CNNs because:
1) GIS cannot evaluate all possible operator set assignments to the two hidden layers (1 and 2). So there are possibly more than 16 operator sets which can yield a better performance than CNN's.
2) If we would not have fixed the operator set of the output layer to 0:{0,0,0}, it is possible to find much more operator assignments to all three layers (2 hidden + 1 output) that may even surpass the performance levels achieved by the top-ranked operator sets, (0,13,0).

*C. Computational Complexity Analysis*

In this section the computational complexity of an ONN is analyzed with respect to the CNN with the same network configuration. There are several factors that affect the computational complexity of an ONN. We shall begin with the complexity analysis of the forward propagation (FP) and then focus on BP.

During FP, the computational difference between an ONN and CNN solely lies in the choice of the operator sets for each neuron/layer. Assuming a unit computational complexity for the operators of CNN (*mul, tanh* and *sum*), Table 5 presents the relative computational complexity factors of the sample operators used in this study. In the worst-case scenario, if *sinc, median* and *tanh* are used as the operator set for each neuron in the network, then a FP in an ONN will be 2.70 x 3.21 x 1 = 8.68 times slower than the one in CNN with the same configuration. In the sample problems addressed in this study, the pool operator determined by 2-pass GIS was always *summation*, therefore, this "worst-case" figure would only be 2.704 times. In practice, we observed a speed deterioration in ONNs usually between 1.4 to 1.9 times with respect to the corresponding FP speed of the (equivalent) CNN. When the configuration CNNx4 was used, ONN's speed became more than twice faster.

**Table 5: Computational complexity factors of each sample nodal, activation and pool operator compared to the operators of CNN (*mul, tanh* and *sum*) during the FP.**

| Nodal (Ψ) | | | | | | Act. (*f*) | Pool (P) |
|---|---|---|---|---|---|---|---|
| *cubic* | *sin* | *exp* | *DoG* | *sinc* | *chirp* | *lin-cut* | *median* |
| 1.01 | 2.21 | 1.78 | 2.55 | 2.70 | 2.41 | 0.99 | 3.21 |

During BP, the computational complexity differences between an ONN and a CNN (having the same configuration) occur at the sub-steps (i-iii) as given in Alg. 1. The 4[th] sub-step (iv) is common for both.

First, the FP during a BP iteration computes all the BP elements as detailed in Alg. 1, i.e., for each neuron, $k$, at each layer, $l$, $f'(x_k^l)$, $\nabla_y \Psi_{ki}^{l+1}, \nabla_{\Psi_{ki}} P_i^{l+1}$ and $\nabla_w \Psi_{ki}^{l+1}$. Only the first BP element, $f'(x_k^l)$, is common with the conventional BP in CNNs, the other three are specific for ONNs and, therefore, cause extra computational cost. Once again when the native operator set of CNNs, 0:{0,0,0} is used, then $\nabla_{\Psi_{ki}} P_i^{l+1} = 1$, $\nabla_y \Psi_{ki}^{l+1} = w_{ik}^{l+1}$ and $\nabla_w \Psi_{ki}^{l+1} = y_k^l$ all of which are fixed and do not need any computation during FP. For the other operators, by assuming again a unit computational complexity for the operators of CNN (*mul, tanh* and *sum*), Table 6 presents the relative computational complexity factors of the sample operators used in this study. In the worst-case scenario, if *sinc, median* and *tanh* are used as the operator set for each neuron in the network, then a FP during a BP iteration will be 1.72 x 8.50 x 1 = 14.63 times slower than the one in CNN with the same configuration. Once again since the pool operator determined by 2-pass GIS for the sample problems addressed in this study, was always *summation*, this "worst-case" figure would be 8.504 times. In practice, we observed a speed deterioration for the FP in each BP iteration usually between 2.1 to 5.9 times with respect to the corresponding speed of the (equivalent) CNN.

**Table 6: Computational complexity factors of each sample nodal, activation and pool operator compared to the operators of CNN (*mul, tanh* and *sum*) during the FP in a BP iteration.**

| Nodal (Ψ) | | | | | | Act. (*f*) | Pool (P) |
|---|---|---|---|---|---|---|---|
| *cubic* | *sin* | *exp* | *DoG* | *sinc* | *chirp* | *lin-cut* | *median* |
| 2.57 | 6.21 | 5.08 | 7.85 | 8.50 | 6.11 | 0.99 | 1.72 |

The 2[nd] step (ii) is actual back-propagation of the error from the output layer to the 1[st] hidden layer to compute delta error of each neuron, $k$, $\Delta_k^l$ at each layer, $l$. This corresponds to the first three BP phases as detailed in Section III.A. The first phase, the delta error computation at the output layer common for ONNs and CNNs, so has the same computational complexity. In the second phase, that is the inter-BP among ONN layers, a direct comparison between Eqs. (24) and (28) will indicate that for each BP step from the next layer to the current layer, i.e., $\Delta y_k^l \xleftarrow{\Sigma} \Delta_i^{l+1}$, the difference lies between the convolution of the next layer delta error, $\Delta_i^{l+1}$ with the static (or fixed kernel) $rot180(w_{ik}^{l+1})$ and the varying (kernel), $\nabla_y P_i^{l+1}$. The latter is obtained by element-wise multiplication of the two derivatives stored in 4D matrices, $\nabla_{\Psi_{ki}} P_i^{l+1}$ and $\nabla_y \Psi_{ki}^{l+1}$, both of which are already computed during the last FP. It is clear that there is no computational complexity difference between the two (fixed vs. varying) convolutions.

The 3[rd] phase, the intra-BP within an ONN neuron is also a common operation with a CNN neuron, and thus has the same computational cost. Finally, for the last phase (or the step (iii) in Alg.1), the computation of the weight and the bias sensitivities, a direct comparison between Eqs. (34) and (38) will indicate that the same computational complexity (as in the 2[nd] phase) exists between the convolution of the next layer delta error, $\Delta_i^{l+1}$ with the static output, $y_k^l$ and the varying (sensitivity), $\nabla_w P_i^{l+1}$. Finally, the 4[th] and last step of the BP given in Alg. 1 is the update of the weights and biases. As a common step, obviously it has also the same computational complexity. Overall, once the BP elements, $f'(x_k^l)$, $\nabla_y \Psi_{ki}^{l+1}, \nabla_{\Psi_{ki}} P_i^{l+1}, \nabla_w \Psi_{ki}^{l+1}$ and $\nabla_y P_i^{l+1} = \nabla_{\Psi_{ki}} P_i^{l+1} \times \nabla_y \Psi_{ki}^{l+1}$ with $\nabla_w P_i^{l+1} = \nabla_{\Psi_{ki}} P_i^{l+1} \times \nabla_w \Psi_{ki}^{l+1}$ are all computed during FP, the rest of the BP phases of ONN will have the same computational complexity as in the corresponding phases for CNNs. So the overall BP speed of ONNs deteriorates due to increased computational complexity of the prior FP. In practice, we observed a speed deterioration for each BP iteration (including FP) usually between 1.5 to 4.7 times with respect to the corresponding speed of a BP iteration in the (equivalent) CNN.

## V. CONCLUSIONS

The ONNs proposed in this study is inspired from two basic facts: 1) bio-neurological systems including the mammalian visual system are based on heterogeneous, non-linear neurons with varying synaptic connections, 2) the corresponding heterogeneous ANN models encapsulating non-linear neurons (aka GOPs) have recently demonstrated such a superior learning performance that cannot be achieved by their conventional linear counterparts (e.g. MLPs) unless significantly deeper and more complex configurations are used [32]-[35]. Empirically speaking, these studies have proven that only the heterogeneous networks with the right operator set and a proper training can truly provide the required kernel transformation to discriminate the classes of a given problem, or to approximate the underlying complex function. In neuro-biology this fact has been revealed as the "neuro-diversity" or more precisely, "the bio-chemical diversity of the synaptic connections". Accordingly, this study has begun from the point where the GOPs have left over and has extended it to design the ONNs in the same way MLPs are extended to realize conventional CNNs. Having the same two restrictions, i.e., "limited connections" and "weight sharing", heterogeneous ONNs can now perform any (linear or non-linear) operation. Our intention is thus to evaluate convolutional vs. operational layers/neurons; hence we excluded the fully-connected layers to focus solely into this objective. Moreover, we have selected very challenging problems while keeping the network configurations compact and shallow, and BP training brief. Further restrictions are applied on ONNs such as a limited operator set library with only 7 nodal and 2 pool operators, and the 2-pass GIS is performed to search for the best operators only for the two hidden layers while keeping the output layer as a convolutional layer. As a result, such a restricted and layerwise homogenous (network-wise heterogeneous) ONN implementation allowed us to evaluate its "baseline" performance against the equivalent and much complex CNNs.

In all problems tackled in this study, ONNs exhibit a superior learning capability against CNNs and the performance gap widens when the severity of the problem increases. For instance, in image denoising, the gap between the average SNR levels in train partition was around 1.5dB (5.59dB vs. 4.1dB). On a harder problem, image syntheses, the gap widens to near 5dB (9.91dB vs. 5.02dB) and on few folds, CNN failed to synthesize the image with a reasonable quality. Finally, on the hardest problem among all, image transformation, the gap exceeded beyond 10dB (10.94dB vs. -0.08dB); in fact, the CNN with the default configuration has failed to transform in all folds. This is also true even though when 4-times more complex CNN model is used and the problem is significantly simplified (only one image transformation rather than 4). This is actually not surprising since a detailed analysis performed during the 2-pass GIS has shown that there are at least 16 other potential ONN models with different operator sets that can perform *better* than the CNN. So for some, relatively easier, problems "linear convolution" for all layers can indeed be a reasonable or even a sub-optimal choice (e.g. object segmentation or even for image denoising) whereas for harder problems, CNNs may entirely fail (e.g. image syntheses and transformation) unless significantly deeper and more complex configurations are used. The problem *therein* lies mainly in the "homogeneity" of the network when the same operator set is used for all neurons/layers. This observation has verified in the 1st fold of the image transformation problem where it sufficed to use a different non-linear operator set *only* for a single layer (layer-2, operator set, 13:{0,1,6}) while all other layers are convolutional. This also shows how crucial it is to find the right operator set for each layer.

There are several ways to improve this "baseline" ONN implementation some of which can be listed as below:
- enriching the operator set library by accommodating other major pool and nodal operators,
- forming layerwise heterogeneous ONNs (e.g. by exploiting [35]) for a superior diversity,
- adapting a progressive formation technique with memory,
- and instead of a greedy-search method such as GIS over a limited set of operators, using a global search methodology which can incrementally design the optimal non-linear operator during the BP iterations.

These will be the topics for our future research.